\documentclass[lettersize,journal]{IEEEtran}
\usepackage{amsmath,amsfonts}
\usepackage{algorithmic}
\usepackage{algorithm}
\usepackage{array}
\usepackage[caption=false,font=normalsize,labelfont=sf,textfont=sf]{subfig}
\usepackage{textcomp}
\usepackage{stfloats}
\usepackage{url}
\usepackage{verbatim}
\usepackage{graphicx}
\usepackage{cite}

\usepackage{bm}
\usepackage{algorithm}
\usepackage{algorithmic}
\usepackage{amsmath}
\usepackage{multirow}
\usepackage{tabu}
\usepackage{booktabs} 
\usepackage[table,xcdraw]{xcolor}
\usepackage{color} 
\usepackage{amsfonts,amssymb} 
\usepackage{ragged2e}

\begin{document}

\title{Improving Feature-based Visual Localization by Geometry-Aided Matching}

\author{ Hailin~Yu,
         Youji~Feng,
         Weicai~Ye,
         Mingxuan~Jiang,
         Hujun~Bao,
         Guofeng~Zhang

        % <-this % stops a space
\thanks{H. Yu, W. Ye, H. Bao and G. Zhang are with the State Key Lab of CAD\&CG, Zhejiang University, and also affiliated with ZJU-SenseTime Joint Lab of 3D Vision. E-mails: hailinyu0414@gmail.com, \{yeweicai, baohujun, zhangguofeng\}@zju.edu.cn. H. Yu is also affiliated with SenseTime Research. G. Zhang is the corresponding author. }% <-this % stops a space
\thanks{Y. Feng and M. Jiang are with SenseTime Research.}}



\maketitle

\begin{abstract}
% Feature matching is a critical step in visual localization, where the accuracy of the camera pose is mainly determined by the established 2D-3D correspondence. A sufficient number of well-distributed 2D-3D correspondences is the key to accurate pose estimation due to the existence of noise. Existing 2D-3D feature matching is typically achieved by finding the nearest neighbors in the feature space, and then removing the outliers by some hand-crafted heuristics. This scheme may lead to a large number of potential real matches being missed or the correct matches that have been found being filtered out, due to the intrinsically limited invariance of local features.
Feature matching is crucial in visual localization, where 2D-3D correspondence plays a major role in determining the accuracy of camera pose. A sufficient number of well-distributed 2D-3D correspondences is essential for accurate pose estimation due to noise. However, existing 2D-3D feature matching methods rely on finding nearest neighbors in the feature space and removing outliers using hand-crafted heuristics, which may lead to potential matches being missed or the correct matches being filtered out. 
In this work, we propose a novel method called Geometry-Aided Matching (GAM), which incorporates both appearance information and geometric context to address this issue and to improve 2D-3D feature matching. GAM can greatly boost the recall of 2D-3D matches while maintaining high precision. We apply GAM to a new hierarchical visual localization pipeline and show that GAM can effectively improve the robustness and accuracy of localization. Extensive experiments show that GAM can find more real matches than hand-crafted heuristics and learning baselines. Our proposed localization method achieves state-of-the-art results on multiple visual localization datasets. Experiments on Cambridge Landmarks dataset show that our method outperforms the existing state-of-the-art methods and is six times faster than the top-performed method. The source code is available at \url{https://github.com/openxrlab/xrlocalization}. 
\end{abstract}


\section{Introduction}\label{sec:introduction}
\IEEEPARstart{V}{isual} localization aims to estimate the 6-Degree-of-Freedom (6DoF) camera pose from a given image or images, which is a fundamental technique for many applications, such as mobile robotics, autonomous driving, and augmented reality.

Feature-based visual localization methods \cite{sarlin2019coarse, sattler2012improving, svarm2016city, liu2017efficient, li2010location, zeisl2015camera, sarlin2018leveraging, taira2018inloc} mainly follow a classic four-stage pipeline: 1) extracting local features (keypoints and descriptors), 2) establishing 2D-3D correspondences by performing feature matching between a query image and an offline reconstructed SfM model, 3) estimating the camera pose by solving a standard PnP \cite{kneip2011novel, lepetit2009epnp} inside a RANSAC \cite{fischler1981random} loop, 4) refining the camera pose with all inliers. In this pipeline, a key step for accurate and robust localization is to search for a sufficient number of correct 2D-3D matches due to the existence of noise and mismatches.

\begin{figure}[tb]
  \centering 
  \includegraphics[width=\columnwidth]{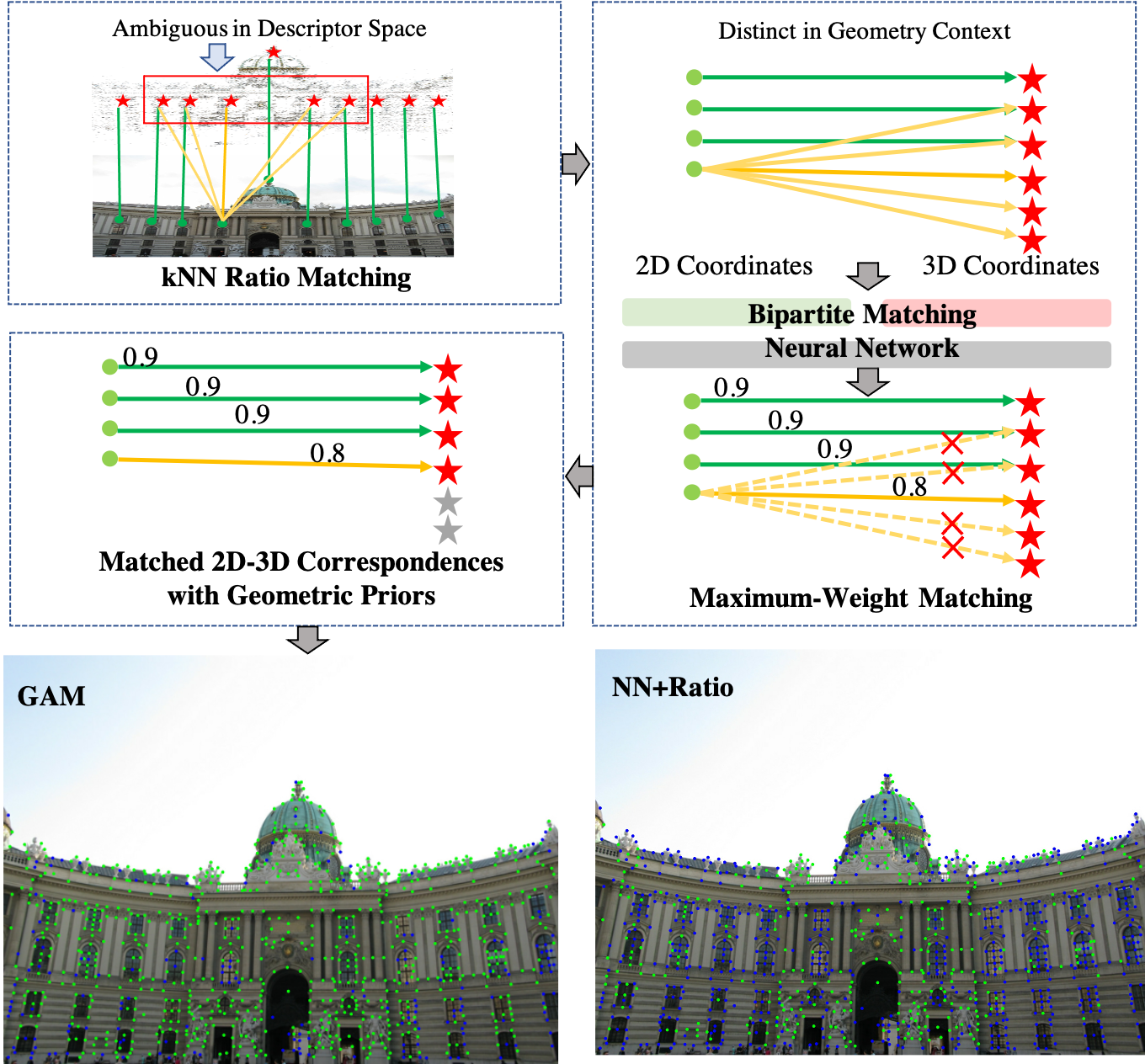}
  \caption{\textbf{Geometry-Aided Matching.} The above part is the illustration of GAM. The below part is two query images matched by GAM and NN matcher followed by ratio test, respectively. Green dots are 2D points that are matched with 3D points correctly. Blue dots are unmatched or erroneously matched 2D points.}
  \label{fig:inlier_compare}
\end{figure}

SIFT \cite{lowe2004distinctive} is the widely used local feature in this pipeline. While exhibiting excellent performance under normal conditions, it is less effective in scenarios with large viewpoint changes or illumination variations\cite{sarlin2019coarse}. To cope with this issue, one possible approach is to strengthen the ability of the local features employed in this pipeline.
Recently, many CNN-based local features have been proposed to extract descriptors solely \cite{luo2019contextdesc, tian2019sosnet, luo2018geodesc} or extract keypoints and descriptors simultaneously \cite{detone2018superpoint, dusmanu2019d2}. These features show impressive robustness against illumination variations and viewpoint changes and perform better than hand-crafted ones on feature matching. Some of them have already been used to improve visual localization \cite{sarlin2019coarse, dusmanu2019d2}.
However, owing to the inborn limited invariance and discriminative capability of local features, some probable real matches may not be discovered by merely searching the nearest neighbor (NN) in the descriptor space.  Some real matches that have been found would be further rejected by ratio test \cite{lowe2004distinctive} that is commonly used to reject ambiguous matches.  This issue is more serious when repetitive patterns exist in scenes.

Recent works \cite{moo2018learning,zhang2019learning} adopt a neural network to learn a function to classify inliers and outliers or to predict matching probability or directly to learn a match function \cite{sarlin2020superglue} for 2D image matching. These data-driven methods show promising results on 2D-2D feature matching. However, the widely used method for 2D-3D matching is still to find the nearest neighbor and followed by ratio test \cite{lowe2004distinctive, sarlin2018leveraging, sarlin2019coarse}, or some heuristic methods \cite{sattler2012improving, liu2017efficient}. 
Some recent methods \cite{sarlin2019coarse, sarlin2020superglue, sun2021loftr, xue2022efficient} perform localization by first performing 2D-2D feature matching between the query image and all retrieved reference images and then lifting to 2D-3D correspondences.  When localizing the query images, multiple 2D-2D feature matching is required as many times as the number of retrieved images. This is very time-consuming and usually takes several seconds to obtain accurate poses. Therefore, this type of approach is not well suited to eliminate tracking drift in AR applications timely.

To address the problem aforementioned, we propose a geometry-aided matching (GAM) method, which directly matches local features in the query image to 3D points in the SfM model. 
% The descriptor of a 3D point can be represented by the mean of all 2D descriptors in the track. 
GAM is a two-step matching method. As shown in Fig.\ref{fig:inlier_compare}, 
% all potential real matches are firstly found by $k$NN ratio matching as candidates, which can recall the correct matches that are not in the nearest neighbor. 
$k$NN ratio matching is able to find the real matches that are not in the nearest neighbor. 
As a price, this brings a large ratio of outliers. For a 2D feature, there could be $k$ incorrect matches. To solve this problem, we turn to leverage the geometric information to find true matches from these candidates, based on the observation that some matches are ambiguous in feature space but distinct in geometry context \cite{moo2018learning, luo2018geodesc}.
Inspired by \cite{moo2018learning}, we propose a  deep neural network to capture the geometric context from candidate matches and then use geometric information to find out true matches.
Unlike \cite{moo2018learning} which treats all 2D-2D correspondences as a 4D point set,  we instead view 2D-3D correspondences as three sets, i.e., 2D point set, 3D point set, and edge sets, then use three networks to process these sets separately. This approach can effectively reuse the features extracted from 2D and 3D point sets. In addition, to find the globally optimal one-to-one matches from many-to-many candidates, we introduce a Hungarian pooling layer to ensure that the output of the network is one-to-one matches, so we call the network bipartite matching neural network (BMNet). Moreover, this pooling layer can effectively avoid the geometrically consistent mismatching problem.

To cope with large scenes, we embed GAM into a hierarchical visual localization pipeline. Based on the retrieved images, a scene retrieval strategy is proposed to expand the retrieval result by exploiting co-visible information provided by the SfM model. This strategy can provide more complete 3D point sets for feature matching. Compared to the preliminary conference version~\cite{yu2020learning}, we make several modifications to improve the robustness. Specifically, we propose a novel $k$NN ratio matching to replace $k$NN matching, which not only improves the quality of matching candidates but also makes it less sensitive to the hyperparameter $k$. We also improve the discriminative power of BMNet by adopting a new approach to generate training data. To summarize,  our major contributions are as follows:
\begin{itemize}
    \item We propose a new 2D-3D matching method GAM that firstly establishes multiple candidate matches for each 2D point depending on visual appearance and then filters incorrect matches depending on geometric context.
    \item We propose a deep neural network BMNet that can deal with many-to-many candidate matches. BMNet can predict the geometric priors for each 2D-3D match and output the globally optimal match set.
    \item We propose a hierarchical visual localization method with a new scene retrieval strategy, which further improves the robustness of pose estimation.
    \item We show that the proposed localization method outperforms the state-of-the-art methods on multiple datasets.
\end{itemize}

The rest of the paper is organized as follows. In Section \ref{related_work}, we discuss the related work. The proposed matching method and neural network are described in Section \ref{gam_matching}. The whole localization pipeline is shown in Section \ref{hloc}.  In Section \ref{exps}, we conduct extensive experiments to justify our proposed approach and then compare it with state-of-the-art methods.

\section{Related Work} \label{related_work}
Visual localization involves many fields, such as local features,  feature matching, and pose estimation. In this section, we review some methods related to our work.

\textbf{Local Feature.} Traditional hand-crafted local features \cite{lowe2004distinctive, rublee2011orb} are sensitive to illumination variation and large viewpoint changes.  When the scene lighting changes, it is difficult to find a sufficient number of correct matches. Recently, many  CNN-based local features \cite{luo2019contextdesc, luo2018geodesc, detone2018superpoint, dusmanu2019d2, tian2019sosnet, bhowmik2020reinforced, pautrat2020online, ng2022ninjadesc} are proposed to replace hand-crafted ones and show a better result on image matching and visual localization \cite{sarlin2019coarse, dusmanu2019d2}. TCDesc \cite{pan2021tcdesc} leverages neighborhood topology consistency to improve local descriptors. 
% However, there are many similar local patterns in real environments, e.g. the facades of modern buildings. Ambiguous matches would be frequently encountered even using powerful learning-based features. The problem can be alleviated by embedding some contextual information from the whole image in the features \cite{luo2019contextdesc, luo2018geodesc}, but this will obscure the raw representation of local details.
However, a large amount of similar local patterns exist in real-world environments, such as the facades of buildings. Even with the use of powerful learning-based features, ambiguous matches would be still unavoidable. This issue can be mitigated by incorporating some global contextual information into the local features \cite{luo2019contextdesc, luo2018geodesc},  but this manner would obscure the local details. GAM fuses global information during the matching stage to address this issue. 

\textbf{Feature Matching.} 
% Feature matching typically takes two steps: 1) establishing initial matches through the Nearest Neighbor (NN) search, 2) filtering ambiguous matches by some strategies, such as ratio test \cite{lowe2004distinctive}, distance threshold, and cross check. In addition to these simple and general strategies,  for the problem of image matching, some works\cite{bian2017gms, ma2019locality} leverage local consistency and use neighborhood consensus to gather correct matches and remove mismatches.
 Feature matching is usually composed of two steps, first searching candidate matches by the Nearest Neighbor (NN) search then removing ambiguous ones by some strategies, such as judging whether being the nearest neighbors to each other (cross check), filtering by descriptor distance threshold, or ratio test \cite{lowe2004distinctive}. In addition to these simple and general strategies,  for the problem of 2D-2D mage matching, some works\cite{bian2017gms, ma2019locality} leverage local consistency to search for more potential matches or remove mismatches.
 % If a geometric model, e.g. a fundamental matrix or an absolute pose, fits for the true matches, robust model estimation approaches can be used to find out the inliers, such as \cite{fischler1981random, chum2005matching, chum2003locally, barath2018graph}.
% Recently, some methods \cite{moo2018learning, zhang2019learning} propose to use a deep neural network to classify initial matches as inliers or outliers, or directly predict the probability of matching \cite{brachmann2019neural}. They have shown impressive results on 2D-2D image matching. \cite{fu2022learning} proposes a matching method that can sense scale differences to cope with large-scale difference image matching.
Recently, some methods \cite{moo2018learning, zhang2019learning} propose to employ a data-driven manner to categorize candidate matches found by nearest neighbor search as inliers or outliers, or directly predict the probability of matching \cite{brachmann2019neural}. They have demonstrated remarkable achievements in 2D-2D image matching. \cite{fu2022learning} proposes a matching method that can sense scale differences to address large-scale difference image matching.
% NCNet\cite{rocco2018neighbourhood} develops an end-to-end trainable convolutional neural network architecture that finds spatially consistent matches by leveraging neighborhood consensus. Feature matching is usually formulated as a quadratic assignment problem that is NP-hard. For a bipartite graph, if the candidate matches and the corresponding weights are given, some methods \cite{crouse2016implementing, kuhn1955hungarian} can find the global optimal solution with the maximum sum of weights. Recent works adopt graph neural networks \cite{chen2021learning, sarlin2020superglue, zhao2021probabilistic, shi2022clustergnn} or a detect-free manner \cite{sun2021loftr, chen2022aspanformer} to improve 2D-2D feature matching.
NCNet\cite{rocco2018neighbourhood} develops an end-to-end trainable convolutional neural network architecture that finds spatially consistent matches by leveraging neighborhood consensus. Feature matching can also be viewed as a quadratic assignment problem. Given a set of initial matches with weights, some methods \cite{crouse2016implementing, kuhn1955hungarian} can find the global optimal matching set. Recently, some works adopt graph neural networks \cite{chen2021learning, sarlin2020superglue, zhao2021probabilistic, shi2022clustergnn} or a detect-free manner \cite{sun2021loftr, chen2022aspanformer} to improve 2D-2D feature matching. 
SuperGlue \cite{sarlin2020superglue}, parallel to our preliminary conference version \cite{yu2020learning}, designs a graph neural network for feature matching and uses self-attention and cross-attention to aggregate global context to achieve robust matching. \cite{shi2022clustergnn} adopts a coarse-to-fine manner to accelerate feature matching. MatchFormer \cite{wang2022matchformer} adopts a detect-and-match scheme to improve image feature matching.  OnePose \cite{sun2022onepose} proposes an architecture based on a graph attention network for 2D-3D feature matching, but it only works on object-level pose estimation and can not apply to large-scale visual localization.

\textbf{Feature-based Localization.} Feature-based visual localization mainly relies on feature matching between a query image and an offline reconstructed map to get 2D-3D correspondences. 
% In the past, many methods \cite{sattler2012improving, svarm2016city, liu2017efficient, li2010location, zeisl2015camera, sattler2015hyperpoints, sattler2011fast} performs direct matching between local features in a query image and the 3D points in an SfM model. As the feature set in the database is typically very large, many previous works aim to improve efficiency. \cite{sattler2011fast} uses a large vocabulary to quantize the local features and prefers to match the features with lower cost in priority. \cite{li2010location} compresses the 3D model and ranks the 3D points according to the chances they might be seen. Direct matching may produce many false correspondences, some methods \cite{liu2017efficient, sattler2015hyperpoints, sattler2012improving} use the constraint of co-visibility to obtain positive matches.  
In the past, many methods \cite{sattler2012improving, svarm2016city, liu2017efficient, li2010location, zeisl2015camera, sattler2015hyperpoints, sattler2011fast} adopt direct matching manner,  which directly performs 2D-3D feature matching between the query image and the whole SfM model.  These works mainly focus on matching efficiency. Active Search \cite{sattler2012improving} uses the local co-occurrence of feature points and combines 2D-to-3D and 3D-to-2D feature matching to achieve excellent registration performance. In addition, some methods \cite{liu2017efficient, sattler2015hyperpoints, sattler2012improving} utilize the constraint of co-visibility to filter out incorrect matches or obtain more correct matches. 
% To cope with large-scale scenarios, some methods adopt a hierarchical paradigm.  Image global features \cite{arandjelovic2016netvlad, torii201524} are firstly extracted to perform image retrieval to find similar images in the database and then match the query features with the points visible in the retrieved images. These methods are more robust if a superior image retrieval method \cite{arandjelovic2016netvlad, wang2022hybrid, tian2021uav, wu2014second} is provided, and easy to integrate with some location prior. Apart from the low-level local features, some approaches \cite{knopp2010avoiding, naseer2017semantics, toft2018semantic} also use high-level semantic information to improve the robustness against seasonal changes or extreme illumination changes.
To handle large-scale scenarios, some methods adopt a hierarchical paradigm. Initially, global features \cite{arandjelovic2016netvlad, torii201524} of images are extracted to perform image retrieval in order to locate similar images in the database. Then, local features are merely matched against these retrieved database images, which effectively reduces the search space of local features. Such methods can be easily combined with some advanced image retrieval or place recognition methods \cite{arandjelovic2016netvlad, wang2022hybrid, tian2021uav, wu2014second}.
Recently, some methods\cite{sarlin2019coarse, sarlin2020superglue, sun2021loftr} get 2D-3D correspondences by performing multiple times 2D-2D feature matching, which is not only time-consuming but required large storage to store all local features. Different from these methods,  our method directly matches the 2D query features with the 3D points, and each 3D point may correspond to only one descriptor. 
Some methods \cite{naseer2017semantics, toft2018semantic} also utilize high-level semantic information for visual localization, which shows good robustness to seasonal changes or illumination variations.

\textbf{End-to-End Localization.}  
We can divide the end-to-end visual localization methods into two categories:  pose regression  \cite{brahmbhatt2018geometry, walch2017image, kendall2015posenet, shavit2021learning, shavit2022camera, chen2022dfnet} and scene coordinates regression \cite{shotton2013scene, brachmann2017dsac, wu2022sc, li2020hierarchical, brachmann2021visual, guan2021scene, zhou2020kfnet, dong2021robust}. They show good robustness in scenarios with illumination variations or non-textured areas. PoseNet \cite{kendall2015posenet} is the first work to train a CNN to directly regress the 6DoF camera pose from a single query image. Many methods \cite{walch2017image, brahmbhatt2018geometry, shavit2021learning, chen2022dfnet} based on this work improve the accuracy by designing new loss functions or novel network architectures. This type of method shows poor localization accuracy and behaves more like image retrieval \cite{sattler2019understanding}. Their accuracy does not meet the requirements of various AR applications. The scene coordinates regression methods \cite{shotton2013scene, brachmann2017dsac, wu2022sc, li2020hierarchical, brachmann2021visual, guan2021scene, zhou2020kfnet, dong2021robust} first regresses the scene coordinates from a query image and then uses a geometric or optimization manner to calculate the camera pose. \cite{shotton2013scene} employs regression forest to infer the coordinates of each pixel in the RGB-D image and then uses RANSAC+PnP to solve the camera pose.  DSAC proposes a differentiable RANSAC to train a scene coordinate regression model in an end-to-end manner. \cite {brachmann2019neural, brachmann2021visual} improves the localization accuracy by regressing dense scene coordinates.  SC-wLS \cite{wu2022sc} adopts a weighted least squares manner for end-to-end scene coordinate regression training. This type of method shows higher accuracy than pose regression. The disadvantage is that they are hard to converge in large-scale scenarios. Both two types of learning-based methods need a large amount of data to train a model as an implicit localization map, which limits its application and scalability.
PixLoc\cite{sarlin2021back} casts camera localization as metric learning, which separates model parameters and scene geometry and shows good generalization to new scenes.  Different from these end-to-end methods,  GAM focuses on 2D-3D feature matching. It is learned separately and not embedded in the whole complicated localization pipeline.

\section{Geometry-Aided Matching}\label{gam_matching}

\begin{figure*}[htb!]
\centering 
\includegraphics[width=0.95\linewidth]{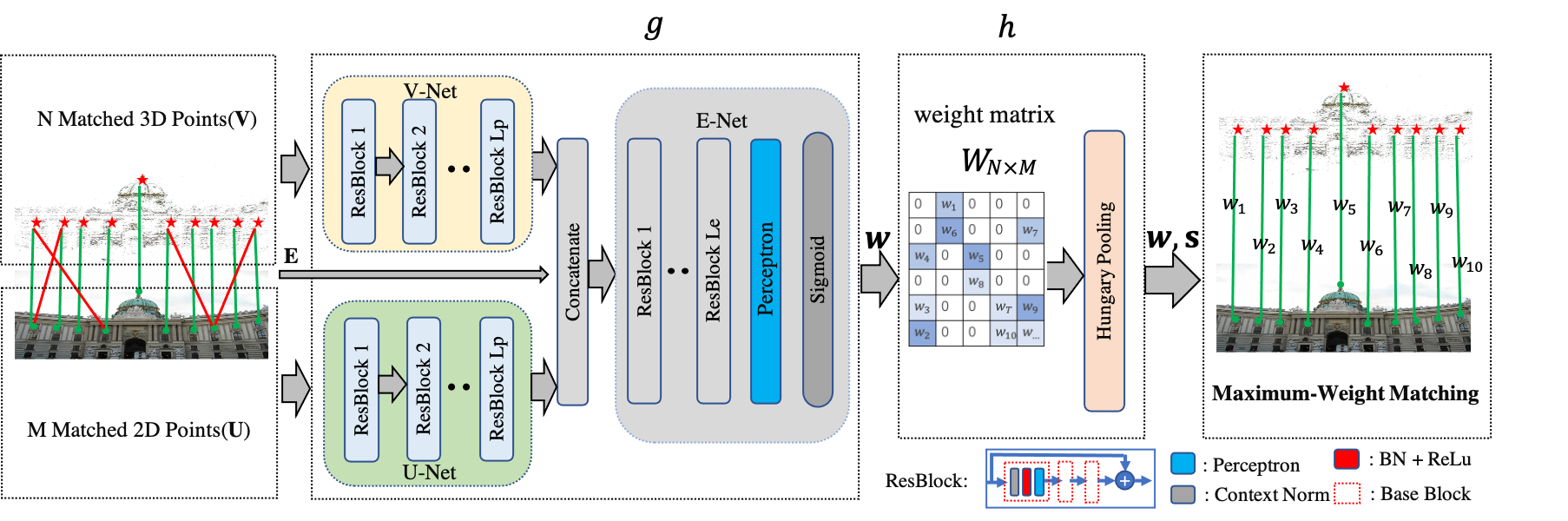}
\caption{\textbf{BMNet Architecture.} 
BMNet is mainly composed of two components $g$ and $h$.  $g$ takes a bipartite graph as input and predicts the probability $w$ for each edge. The bipartite graph consists of a 2D point set ($\bm{\mathcal{U}}$), a 3D point set ($\bm{\mathcal{V}}$), and a 2D-3D match set ($\bm{\mathcal{E}}$). Correct matches in the match set are depicted in green and incorrect ones are in red. $h$ creates an M by N weight matrix $\bm{W}$, filling the probability of each edge to the matrix, and finding the maximum weight matching. }
% BMNet takes a bipartite graph as input, predicting the probability of each edge and outputting the maximum-weight matching. BMNet takes a bipartite graph as input and outputs the maximum-weight matching and the corresponding probability of being an inlier for each selected correspondence. The input graph is composed of the 2D point set ($\bm{\mathcal{U}}$), the 3D point set ($\bm{\mathcal{V}}$), and the 2D-3D correspondence set ($\bm{\mathcal{E}}$) in which the inliers are displayed in green and the outliers are displayed in red. Three sub-networks (U-Net, V-Net, and E-Net) are used to extract the geometric features for $\bm{\mathcal{U}}$, $\bm{\mathcal{V}}$, and $\bm{\mathcal{E}}$ respectively.}
\label{fig:gpnet}
\end{figure*}

In this section, we describe the proposed 2D-3D matching method GAM. We first present the problem formulation, describe BMNet architecture, and then describe how to train BMNet. The details are shown below.

\subsection{Problem Formulation}
Given two feature sets, $\bm{\mathcal{A}}$ and $\bm{\mathcal{B}}$, $\bm{\mathcal{A}}=\{\bm{f}_1^a, ..., \bm{f}_M^a\}$ is 2D feature set and $\bm{\mathcal{B}}=\{\bm{f}_1^b, ..., \bm{f}_N^b\}$ is 3D feature set. We denote feature as $\bm{f}_i = (\bm{p}_i, \bm{d}_i)$, where $\bm{p}_i$ is coordinate $(x_i, y_i)$ for the 2D feature set $\bm{\mathcal{A}}$ and $(X_i, Y_i, Z_i)$ for the 3D feature set $\bm{\mathcal{B}}$. $\bm{d}_i \in \mathbb{R}^D $ is the descriptor that can be extracted using SuperPoint \cite{detone2018superpoint} or other local features such as D2Net \cite{dusmanu2019d2}. Each 3D point descriptor $\bm{d}_i$ is represented by the mean of all 2D descriptors in the track.

We first introduce $k$NN ratio matching to establish 2D-3D correspondences depending on similar visual appearance. Specifically, for 2D feature $\bm{f}_i^a$, $K$ nearest 3D points in descriptor space are firstly found as initial matches $\{\bm{m}_{i1}, ..., \bm{m}_{iK}\}$, which distances are denoted as $d_{i1} , ... , d_{iK} $ and satisfy $d_{i1} < ... < d_{iK} $. Then, the matches that satisfy $d_{i1} / d_{ik} \geq r$ are selected as candidates, where $k$ is from $2$ to $K$ and $r$ is the ratio threshold. If $r=1$, it equals finding the nearest neighbor, and if $r=0$, it equals finding the $K$ nearest neighbors.  

BMNet is then introduced to filter outliers from candidate matches by leveraging geometric context. We view these candidate matches and corresponding 2D and 3D points as a bipartite graph denoted as $\bm{\mathcal{G}} = (\bm{\mathcal{U}}, \bm{\mathcal{V}}, \bm{\mathcal{E}})$. $\bm{\mathcal{U}} = \{\bm{p}_1^a, ..., \bm{p}_{M}^a\}$ is the 2D point set that constructed by all matched 2D points.  $\bm{\mathcal{V}} = \{\bm{p}_1^b, ..., \bm{p}_{N}^b\}$ is the 3D point set that constructed by all matched 3D points. $\bm{\mathcal{E}} = \{\bm{e}_1, \bm{e}_2, ..., \bm{e}_T\}$ is the edge set, where each edge $\bm{e}_k = (i, j)$ with $1\leq k \leq T$ represents that there is a match between the $i$-th 2D point and the $j$-th 3D point. $T$ is the number of all candidate matches. 
In order to find the maximum-weight matching $\bm{\mathcal{M}}$, BMNet first predicts a weight $w_k$ for each edge $\bm{e}_k$.
% which is a subset of $\bm{\mathcal{E}}$. 
Here, the weight $w_k$ represents the probability that $\bm{e}_k$ is an inlier. Then, BMNet finds the maximum weight matching depending on the initial bipartite graph and the predicted weights.
% BMNet aims to predict a weight $w_k$ for each edge $\bm{e}_k$ and then find the maximum-weight matching $\bm{\mathcal{M}}$, which is a subset of $\bm{\mathcal{E}}$. The weight $w_k$ represents the likelihood that $\bm{e}_k$ is an inlier. 
% The outputs of BMNet are formally written as a weighting vector $\bm{w} = (w_1, w_2, ..., w_T)$ with $w_k \in [0, 1]$ and an assignment vector $\bm{s} = (s_1, s_2, ..., s_T)$ with $s_k \in \{0, 1\}$ which indicates whether the $k$-th edge is contained in the maximum-weight matching $\bm{\mathcal{M}}$.
The outputs of BMNet are formally represented as a weighting vector $\bm{w} = (w_1, w_2, ..., w_T)$ with $w_k \in [0, 1]$ and an assignment vector $\bm{s} = (s_1, s_2, ..., s_T)$ with $s_k \in \{0, 1\}$, denoting whether the $k$-th edge is included in the maximum-weight matching $\bm{\mathcal{M}}$.

Finally, GAM selects the edges with $s_k = 1$ 
% and $w_k > p$
as final matches, which are used to solve camera pose.

\subsection{Network Architecture}\label{sec:netarc}
% This part describes the proposed deep architecture BMNet.  The overall architecture is illustrated in Fig.\ref{fig:gpnet}. The input is a bipartite graph $\bm{\mathcal{G}}$ constructed from the 2D-3D correspondences established by $k$NN ratio matching.  The workflow of BMNet can be divided into two steps. The first step is to predict weighing vector $\bm{w} = g(\bm{\mathcal{G}}; \bm{\theta}) $, where $\bm{\theta}$ is the learning parameters. The second step is to find out the maximum-weight matching result from  $\bm{\mathcal{G}}$ and $\bm{w}$, denoted as $\bm{s} = h(\bm{\mathcal{G}}, \bm{w})$, where the function $h$ is non-parametric. The final output is obtained by combining the above two parts, denoted as:
This part describes the proposed deep architecture BMNet. As illustrated in Fig.\ref{fig:gpnet}, the input is a bipartite graph $\bm{\mathcal{G}}$ constructed from the 2D-3D correspondences established by $k$NN ratio matching. BMNet is comprised of two parts arranged in a sequence. Firstly, it predicts a weighing vector $\bm{w} = g(\bm{\mathcal{G}}; \bm{\theta})$, where $\bm{\theta}$ denotes the learning parameters. Secondly, it finds out the maximum-weight matching from $\bm{\mathcal{G}}$ and $\bm{w}$, denoted as $\bm{s} = h(\bm{\mathcal{G}}, \bm{w})$. Here, the function $h$ is non-parametric. The final output is obtained by combining the above two parts, denoted as:
\begin{equation}
    (\bm{w}, \bm{s}) = f_{\bm{\theta}}(\bm{\mathcal{G}})
\end{equation}

% Since the three parts, i.e. the two point sets $\bm{\mathcal{U}}$ and $\bm{\mathcal{V}}$, and the edge set $\bm{\mathcal{E}}$, of the bipartite graph, are all unordered sets, we use Perceptron \cite{hastie2009elements} as the basic layers to extract the geometric features and use Context Normalization \cite{moo2018learning} to aggregate the global information. 

% We use MLP\cite{hastie2009elements} as the basic layers to extract the geometric features and use Context Normalization \cite{moo2018learning} to aggregate the global information. 

% As shown in Fig.\ref{fig:gpnet}, $g(\bm{\mathcal{G}}; \bm{\theta})$ contains three sub-networks, represented by U-Net, V-Net, and E-Net, respectively. The input to U-Net is the 2D point set $\bm{\mathcal{U}}$, which can be represented as a  $M \times 2$ matrix. U-Net embeds the $M$ 2D point coordinates into $M$ $d_1$-dimensional vectors, so the output of U-Net is a $M \times d_1$ matrix $\bm{X_u}$. Similarly, the input to V-Net is the 3D point set $\bm{\mathcal{V}}$, which can be represented as a $N \times 3$ matrix. V-Net embeds the $N$ 3D point coordinates into $N$ $d_2$-dimensional vectors, and the output of V-Net is a $N \times d_2$ matrix $\bm{X_v}$.
% As depicted in Fig.\ref{fig:gpnet}, 
$g(\bm{\mathcal{G}}; \bm{\theta})$ contains three sub-networks, namely U-Net, V-Net, and E-Net. The input to U-Net is the 2D point set $\bm{\mathcal{U}}$, which is represented as a $M \times 2$ matrix. U-Net embeds the $M$ 2D point coordinates into $M$ $d_1$-dimensional vectors, resulting in a $M \times d_1$ matrix $\bm{X_u}$. Similarly, the input to V-Net is the 3D point set $\bm{\mathcal{V}}$, which is represented as a $N \times 3$ matrix. V-Net embeds the $N$ 3D point coordinates into $N$ $d_2$-dimensional vectors, generating a $N \times d_2$ matrix $\bm{X_v}$. The input to E-Net is generated by concatenating $\bm{X_u}$  and $\bm{X_v}$ according to the edge set $\bm{\mathcal{E}}$. Specifically, for an edge $\bm{e}_k = (i, j)$, the feature vector input to E-Net can be obtained as follows: 
\begin{equation}
    \bm{X_e}^k = \left[ \bm{X_u}^i || \bm{X_v}^j\right],
\end{equation}
where $\bm{X_*}^r$ denotes the $r$-th row of matrix $\bm{X_*}$ and $\left[\cdot||\cdot\right]$ denotes concatenation. For all edges in set $\bm{\mathcal{E}}$,  a $T \times (d_1+d_2)$ matrix is obtained as the input to E-Net. Here, both $d_1$ and $d_2$ are set to $128$. The output of E-Net is a $T$-dimensional vector. A sigmoid layer is placed on the top of E-Net in order to ensure that each element of the output vector $\bm{w}$ is within the range of $[0, 1]$. Each element represents the probability that the corresponding edge is an inlier. 

Here, every sub-network is stacked by the base block which is composed of perceptrons, Batch Normalization (BN) \cite{ioffe2015batch}, and Context Normalization (CN) \cite{moo2018learning}. 

\begin{figure}[htb!]
\centering 
\includegraphics[width=1\linewidth]{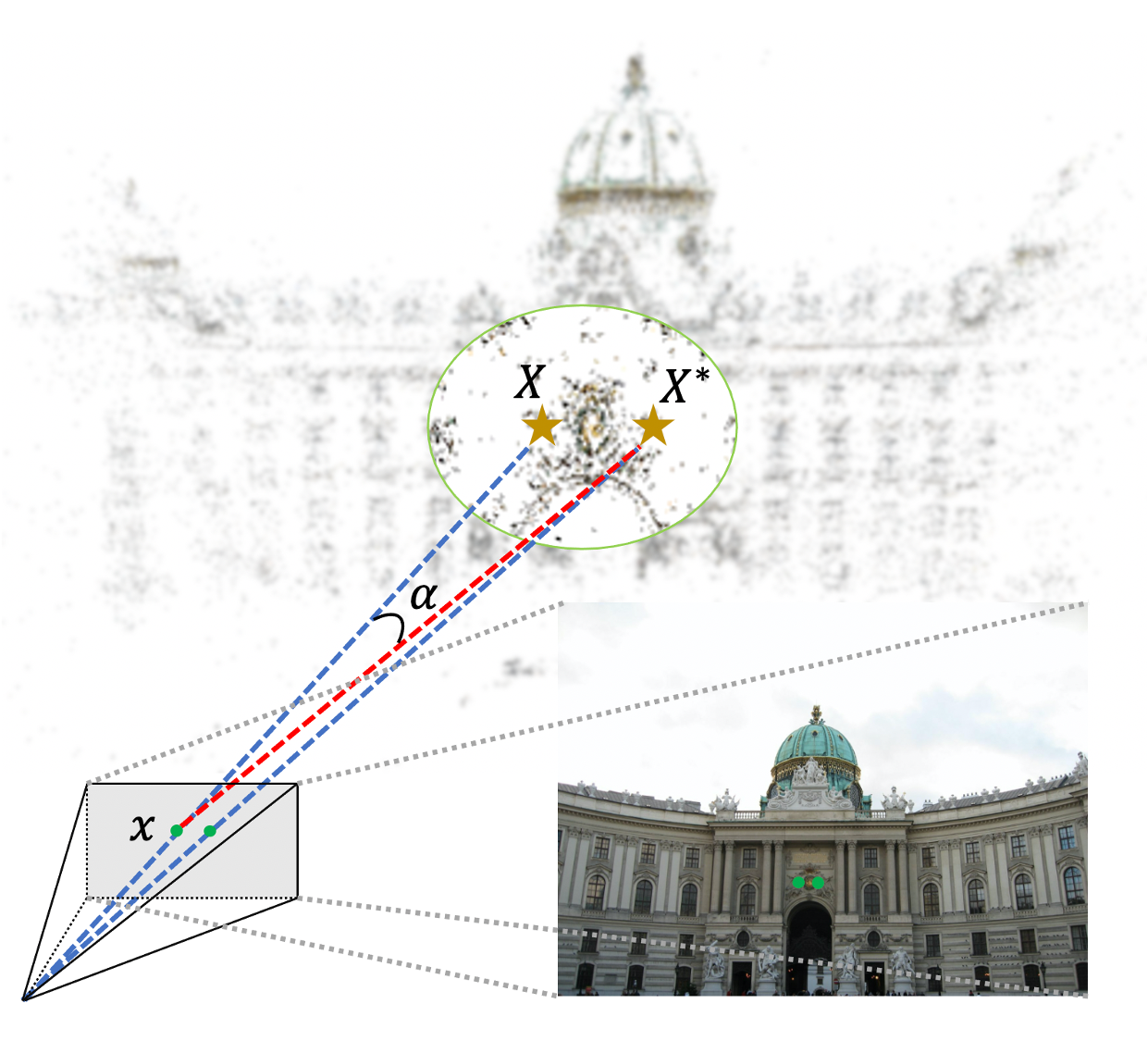}
\caption{
% $\bm{X}$ and $\bm{X^*}$ are two 3D points, and $\bm{x}$ is the projection of $\bm{X}$ on the image plane. The correspondence $(\bm{X^*}, \bm{x})$ is an outlier established by feature matching. $(\bm{X}, \bm{x})$ is the correct match. When the angle $\alpha$ is small, two correspondences $(\bm{X^*}, \bm{x})$ and  $(\bm{X}, \bm{x})$ should have similar weights from a geometric view, but from a learning view, the weight of correspondence $(\bm{X^*}, \bm{x})$ should be much smaller than correspondences $(\bm{X}, \bm{x})$ due to their respective label. This produces a conflict that makes the network hard to train.
$\bm{X}$ and $\bm{X^*}$ are two 3D points that are look similar. $\bm{x}$ is a detected keypoint whose corresponding 3D point is  $\bm{X}$. Due to $\bm{X}$ and $\bm{X^*}$ have similar local patterns,  both $(\bm{X^*}, \bm{x})$ and $(\bm{X}, \bm{x})$  are considered as matches at feature matching.  When the angle $\alpha$ is small, two matches $(\bm{X^*}, \bm{x})$ and  $(\bm{X}, \bm{x})$ should have similar weights from a geometric view.  However, the match $(\bm{X}, \bm{x})$  should be expected to get a higher weight than the match $(\bm{X^*}, \bm{x})$, which is because the former is an inlier and the latter is an outlier. This produces a conflict that makes the network hard to train.
}
\label{fig:proj}
\end{figure}

\textbf{Hungarian Pooling.} 
% If we train  $g(\bm{\mathcal{G}}; \bm{\theta})$  directly, the network parameters will be very difficult to learn because the geometric consistency may conflict with the supervision. 
If directly training $g(\bm{\mathcal{G}}; \bm{\theta})$, the network learning would be disturbed by the potential conflict between the geometric consistency and the supervision.
The conflict is shown in Fig.\ref{fig:proj}.  
% A 2D image point is matched with two 3D points being close in 3D space.
 A 2D image point is matched with two 3D points, and the angle constructed by two correspondences is small.
These two 2D-3D matches may show almost the same reprojection errors. At this time,  the network is liable to extract the same geometric features and predict similar weights for them. These two correspondences are considered geometric consistent. 
% However, only one in the two correspondences may be an inlier, and the rest an outlier. This discrepancy that the multiple correspondences have similar geometric features but with different labels makes the network hard to converge.
However, only one of the two correspondences may be an inlier, and the other is an outlier. This produces a conflict that the multiple matches have almost the same geometric features but belong to different categories, which makes the network hard to learn.

% To solve this problem, we introduce the Hungarian algorithm \cite{kuhn1955hungarian} into the network for end-to-end training.  Hungarian algorithm can find the global optimal one-to-one matches. Because only one of the two correspondences is selected, the discrepancy between the geometric consistency and the supervision can be eliminated.
To deal with this issue, we introduce a novel layer termed Hungarian pooling, which is embedded with the Hungarian algorithm \cite{kuhn1955hungarian} to determine the maximum weight matching from the original many-to-many candidate matches. The maximum weight matching is a subset of the original match set and is ensured to be one-to-one, which can naturally eliminate the conflict mentioned above.

\begin{figure*}[htb!]
\centering 
\includegraphics[width=1.0\linewidth]{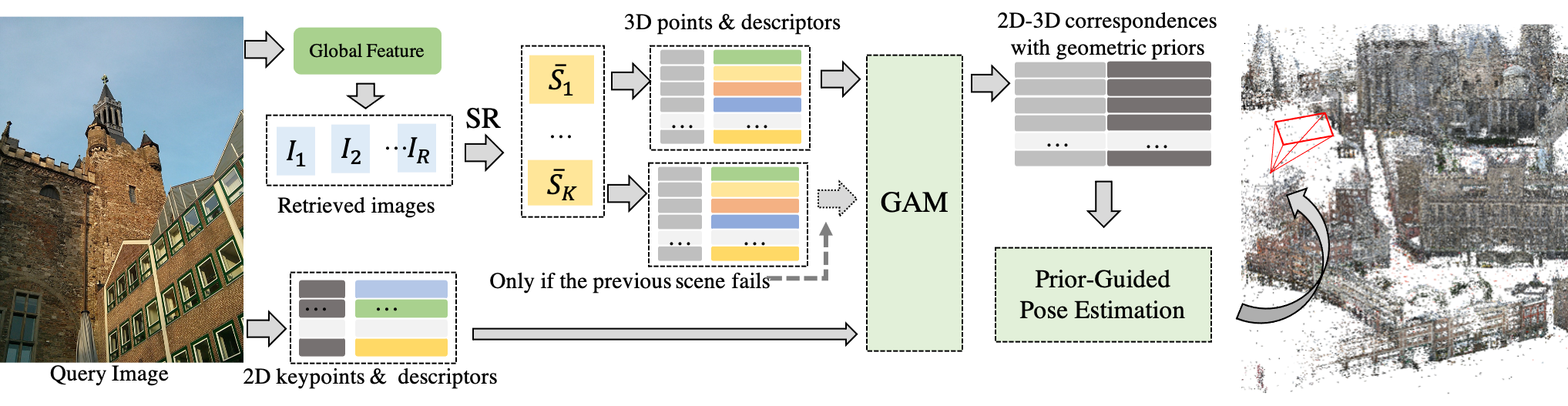}
\caption{\textbf{Hierarchical Visual Localization Pipeline.} GAM is embedded in a hierarchical visual localization pipeline with a scene retrieval strategy. Global and local features are extracted from a query image and are used for scene retrieval and 2D-3D feature matching (GAM). Finally, prior-guided pose estimation is performed depending on the output of GAM.}
\label{fig:pipeline}
\end{figure*}

% The HungrBased on the weight vector $\bm{w}$ predicted by $g(\bm{\mathcal{G}}; \bm{\theta})$  and the bipartite graph $\bm{\mathcal{G}}$, a weight matrix $\bm{W}$ is constructed as:  
The Hungarian pooling layer is placed at the end of BMNet. The input 
 to Hungarian pooling is an $M$ by $N$ weight matrix $\bm{W}$.  All the values in the weight matrix $\bm{W}$ are initialized to zero. Then fill in the probability of all matches predicted by $g(\bm{\mathcal{G}}; \bm{\theta})$, which can be formulated as the following: 
\begin{equation}
  \bm{W}[\bm{e}_k[0], \bm{e}_k[1]] \leftarrow w_k,
\end{equation}
% where the unfilled elements of $\bm{W}$ is set to 0. Then the Hungarian algorithm is applied on this weight matrix $\bm{W}$ to get the maximum-weight matching $\bm{\mathcal{M}}$. The assignment vector $\bm{s}$ is obtained by
The weight matrix $\bm{W}$ is then fed into the Hungarian pooling layer to output an assignment vector $\bm{s}$. The length of the assignment vector  $\bm{s}$ is equal to the size of the edge set.   The elements in $\bm{s}$ indicate whether the corresponding edge (or match) is in the maximum weight matching $\bm{\mathcal{M}}$, which can be formulated as the following:
\begin{equation}
  s_k =\left\{
    \begin{aligned}
       1 && \bm{e}_k \in \bm{\mathcal{M}}; \\
       0 && otherwise.
    \end{aligned}
  \right.
\end{equation}

% Since the output edges come from a subset of the input edges, the layer introducing the Hungarian algorithm can be regarded as a special sampling layer, which we referred to as the Hungarian pooling. The back-propagation used in the end-to-end training is formulated as: 
This formulation can be seen as a special sampling, so we call it Hungarian pooling, which can be embedded into the network and trained in an end-to-end manner.  The back-propagation of this layer is formulated as: 
\begin{equation}
   \frac{\partial h(\bm{\mathcal{G}}, \bm{w})}{\partial w_k} = \left\{
    \begin{aligned}
       1 && \bm{e_k} \in \bm{\mathcal{M}}; \\
       0 && otherwise.
    \end{aligned}
  \right.
\end{equation}

\subsection{Learning from SfM Model}
In this section, we describe how to learn the parameters of BMNet. There are three parts: 1) training data generation, 2) data augmentation, and 3) the loss function. 

\textbf{Training Data Generation.} The training data can be generated automatically using the SfM technique. We generate a bipartite graph for each image that has been registered by SfM.  Specifically, for an image $\bm{I}$  in the SfM model, we take the extracted 2D keypoints and their corresponding descriptors to form the 2D feature set. Then,  all the 3D  points observed by the images that are co-visible with image $\bm{I}$  and its corresponding descriptors are fetched to form the  3D  feature set. The descriptor of each 3D point is represented as the mean of all 2D descriptors in the track except in image $\bm{I}$.  The edge set is generated online at the training stage.

\textbf{Negative Sample Mining.}  Different from the preliminary conference work \cite{yu2020learning} where all negative edges are generated in a completely random manner, we instead perform $k$NN ratio matching online with a fixed number of randomly selected features from both 2D and 3D feature sets to generate positive and negative edge samples during training. The negative samples generated in this way tend to be harder than that of random selection while being consistent with the inference stage in terms of data distribution.  This can effectively improve the discriminative power of the training model.

\textbf{Loss Function.} Finding true matches from the candidates is essentially a classification problem, so we use the widely used cross-entropy loss function for training:
\begin{equation}
  \mathcal{L}_{\bm{\theta}} = \frac{1}{T}\sum^{T}_{k=1}(t_klog(w_k)s_k+(1-t_k)log(1-w_k)s_k).
\end{equation}
where $t_k = 1$ is for true match otherwise $t_k = 0$.

\section{Hierarchical Visual Localization} \label{hloc}
In this section, we embed the proposed geometry-aided matching method (GAM) into a hierarchical visual localization pipeline, shown in Fig.\ref{fig:pipeline}. For a query image, its global feature and local features are extracted. The global feature is used for coarse localization to determine the 3D point set to be matched. Both the extracted local features and the 3D point set combining corresponding descriptors are fed into GAM to get global optimal 2D-3D correspondences. The whole localization process is divided into three modules namely scene retrieval, 2D-3D feature matching, and prior-guided pose estimation. The following describes these modules in detail.

\textbf{Scene Retrieval.} 
% We define a set of the 3D points observed in one image in an SfM model as a meta scene, so we can get a set of the meta scenes $\bm{\mathcal{S}} = \{S_1, S_2, ..., S_N\}$ from a given SfM model, where $N$ is the number of registered images. We use the global descriptor of the query image to retrieve the top $R$ images $\bm{\mathcal{I}} = \{I_1, I_2, ..., I_R\}$ from the database. The corresponding meta scenes are denoted as $\bm{\hat{\mathcal{S}}} = \{\hat{S}_1, \hat{S}_2, ..., \hat{S}_R\}$.  Instead of directly using the meta scenes for feature matching, we further perform an expansion. We denote $\beta = \left| S_i \cap S_j \right|$, which is the number of co-visible 3D points of two scenes.  $\beta > 0$ means that the meta scenes $S_i$ and $S_j$ are co-visible. For each retrieved meta scene $\hat{S_i}$ in $\bm{\hat{\mathcal{S}}}$, we expand it according to co-visibility. This is achieved by finding all meta scenes from $\bm{\mathcal{S}}$ that are co-visible with $\hat{S_i}$ and then selecting the top $m$ ones with the most co-visible points. Then, all the selected $m$ meta scenes are merged as an expanded scene $\bar{S}_k$.  The expansion is performed for the retrieved meta scenes in descending order according to the retrieval scores. If the next retrieved meta scene has already appeared in the previous scenes, we simply skip it. Finally, we get a set of scenes $\bm{\bar{\mathcal{S}}}=\{\bar{S}_1, \bar{S}_2, ..., \bar{S}_K\}$, which will be used for local feature matching.
 We define a set of 3D points observed by a reference image in an SfM model as a meta scene. In this way,  we can obtain a set of meta scenes $\bm{\mathcal{S}} = \{S_1, S_2, ..., S_N\}$ from the given SfM model, $N$ represents the number of reference images. We retrieve the top $R$ reference images $\bm{\mathcal{I}} = \{I_1, I_2, ..., I_R\}$ from the database according to the image global descriptor, and the corresponding meta scenes are denoted as $\bm{\hat{\mathcal{S}}} = \{\hat{S}_1, \hat{S}_2, ..., \hat{S}_R\}$.  Instead of directly performing feature matching between these retrieved meta scenes and the query image,  
 % we first expand each meta scene to incorporate more possibly visible meta scenes.  
 we first expand each meta scene depending on co-visible information provided by the SfM model.  
 We denote the number of co-visible 3D points of two scenes as  $\beta = \left| S_i \cap S_j \right|$.  $\beta > 0$ indicates that two meta scenes $S_i$ and $S_j$ are co-visible.  For a meta scene $\hat{S_i}$, the expansion is achieved by searching top $m$ meta scenes with the most co-visible points from $\bm{\mathcal{S}}$. Then, these $m$ meta scenes are merged as an expanded scene $\bar{S}_k$.  If $\hat{S_i}$ has already appeared in the previous expanded scenes, it can be skipped to the next retrieved meta scene. Finally, we obtain a set of expanded scenes $\bm{\bar{\mathcal{S}}}=\{\bar{S}_1, \bar{S}_2, ..., \bar{S}_K\}$.

\textbf{2D-3D Feature Matching.} GAM is performed sequentially according to the order of the scene retrieval and outputs matched 2D-3D correspondences. For the $k$-th expanded scene $\bar{S}_k$,  we fetch its 3D points and corresponding descriptors to construct the 3D feature set.  Local features are extracted from the query image to construct the 2D feature set.   Then,  both two feature sets are fed into GAM and output the maximum weight matching $\bm{\mathcal{M}}$. We perform $k$NN ratio matching depending on Euclidean distances, which can be achieved efficiently through matrix operations when descriptors are normalized.

\textbf{Prior-Guided Pose Estimation.} 
% Since there are still some false matches in $\bm{\mathcal{M}}$, we apply a PnP solver inside the RANSAC loop. In the RANSAC loop, the probability of sampling 2D-3D correspondences is decided by the likelihood predicted by BMNet. This allows us to sample possible inliers with larger chances. 
Since there are still some erroneous matches in $\bm{\mathcal{M}}$, we employ a PnP solver within the RANSAC loop, in which the probability of selecting 2D-3D correspondences is determined by the weights predicted by BMNet. This enables the RANSAC algorithm to sample potential inliers with higher probabilities.

% Our pipeline is partially inspired by the current state-of-the-art method \cite{sarlin2019coarse}. It clusters the retrieved images to form several scenes by using co-visibility. Fine localization is performed against each scene in descending order according to the size of the scene. The localization is terminated if a credible result, e.g. the number of inliers is larger than a threshold, is obtained. 
The proposed localization method is partially based on the state-of-the-art approach\cite{sarlin2019coarse}. It clusters the retrieved images to form multiple scenes by exploiting the co-visibility provided by the sparse SfM model. Feature matching is then conducted on each scene in descending order according to the scene's size. The algorithm is terminated when a reliable localization result, such as the number of inliers exceeding a given threshold, is obtained.
% We use the same early-stop strategy, but the construction of the scene is different. First, the scene in \cite{sarlin2019coarse} contains the retrieved images only, we make an expansion in the original SfM model to effectively find more relevant 3D points.
In our method, we modify the usage of retrieved images. Instead of directly clustering the retrieved images based on co-visibility, we employ the co-visibility information to make an expansion to obtain more 3D points that may be visible in the query image.
Furthermore, the clustering manner in \cite{sarlin2019coarse} causes an unlimited transitive co-visibility. For instance, two images without any co-visible points can be clustered together as a single cluster if a third image is co-visible to both of them separately. This will result in the cluster having too large a span in space, which is not conducive to feature matching and data normalization. 
% In the work, we limit the size of the scene and prefer the non-transitive co-visibility to get a cleaner scene. 
In our work, we restrict the size of the scene and employ non-transitive co-visibility to obtain a more organized scene.

The proposed localization method is not limited to a specific image retrieval method and a specific local feature. In this work, if not otherwise specified, we use NetVLAD \cite{arandjelovic2016netvlad} and SuperPoint\cite{detone2018superpoint} as the global and local feature extractors, respectively.

\section{Experiments}\label{exps}
In this section, we first verify the effectiveness of the proposed 2D-3D matching method GAM through extensive experiments and demonstrate the state-of-the-art localization performance on multiple public datasets. We first discuss the architecture details and our training configuration. Then, we evaluate GAM and the proposed localization method. Finally, we compare our method with state-of-the-art methods and provide an ablation study.

\subsection{Architecture Details and Training Setup}
The number of blocks of U-Net and V-Net $L_p$ and E-Net $L_e$ is set to 5 and 18, respectively. The model is implemented in PyTorch \cite{paszke2019pytorch}. We use the MegaDepth dataset \cite{li2018megadepth}, which includes 196 different locations. Each location provides an SfM model reconstructed by COLMAP \cite{schonberger2016structure} using SIFT. We select four locations, which include about 12k images. We fix all image poses and use HLoc toolbox\cite{sarlin2020hloc, sarlin2020superglue} to reconstruct the selected 4 locations with SuperPoint. We use the reconstructed SfM models to construct our training set. If not otherwise specified, all learned methods used in the following experiments are trained under this training set. BMNet used in GAM is trained using an SGD optimizer with an initial learning rate $0.001$ and batch size $1$. It converges after $140$ epochs of training on one GTX1080Ti GPU. The following experiments, if not otherwise specified, use GAM with $k=3$ and $ratio=0.7$ as the default configuration.

\begin{figure*}[htb!]
  \centering
  \includegraphics[width=0.98\linewidth]{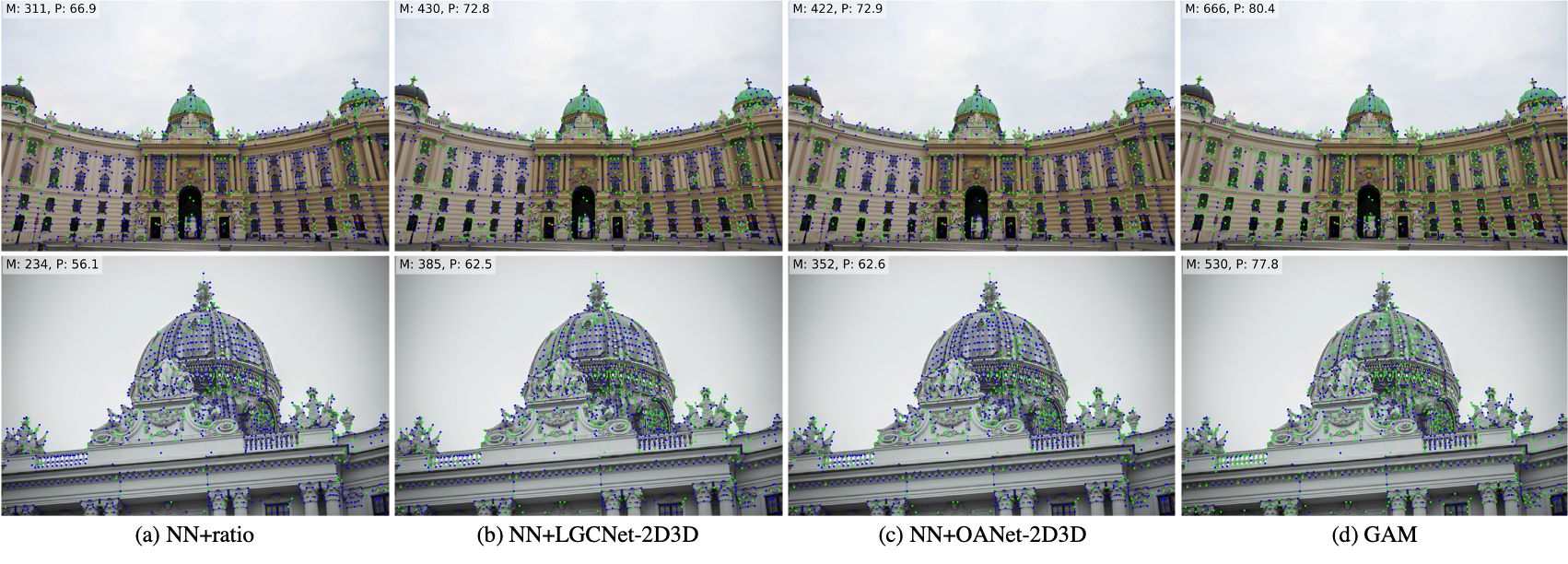}
  \caption{\textbf{Qualitative 2D-3D matching}. Blue dots and green dots represent unmatched keypoints and truly matched keypoints, respectively. M and P represent the number of correct matches and the match precision, respectively.}
  \label{fig:2D3Dmatching}
\end{figure*}

\subsection{2D-3D Matching}\label{sec:exp_matching}
We first perform 2D-3D matching experiments and show that GAM can find more true matches with higher precision than hand-crafted methods and learning methods. 

\textbf{Dataset.} We choose one location from MegaDepth dataset \cite{li2018megadepth} as the test dataset for 2D-3D matching evaluation. The chosen location is totally different from the training set. The dataset includes 508 images. We fix all image poses and use HLoc toolbox \cite{sarlin2020hloc, sarlin2020superglue} to reconstruct this location with SuperPoint \cite{detone2018superpoint}. 

\textbf{Metrics.} We report match precision and match recall, which are computed from the ground truth 2D-3D correspondences. Match precision refers to the ratio of ground truth correspondences being matched over the number of all matches. Match recall refers to the ratio of true correspondences being matched over the number of all true correspondences. We further use matched 2D-3D correspondences to compute the poses for all test images and report the median positional errors (MPE, no scale) and the median rotation errors (MRE, $^\circ$) for evaluating of the accuracy of estimated poses.

\textbf{Compared with Baselines.} We compare GAM with both traditional hand-crafted methods and learned matching methods. All methods use SuperPoint feature. For all baselines, we use the Nearest Neighbor (NN) matcher, finding the nearest neighbor from all 3D points for each 2D point in the descriptor space. The hand-crafted outlier rejection methods include ratio test (ratio), distance threshold (distance), and being the nearest neighbors to each other (cross check). 
% To the best of our knowledge, ratio test is commonly the most popular outlier rejection method for 2D-3D matching and there is no learning-based method directly applied to 2D-3D matching. 
OnePose \cite{sun2022onepose} only works on object-level 2D-3D feature matching and can not handle feature matching in such large-scale scenes. Except for OnePose, there is no other learning-based method directly used for 2D-3D feature matching. 
Therefore, we migrate some of the recently proposed methods that perform well in 2D-2D matching to 2D-3D for comparison. We replace the input of LGCNet \cite{moo2018learning} and OANet \cite{zhang2019learning} from 2D-2D correspondences to 2D-3D correspondences, namely LGCNet-2D3D and OANet-2D3D respectively. The training method is kept the same as BMNet. We try to train SuperGlue \cite{sarlin2020superglue} for 2D-3D matching but it cannot be applied to this scale of matching due to the limitation of GPU memory.

\begin{table}[htb!]
  \setlength{\tabcolsep}{2mm}
  \centering %
  \caption{2D-3D feature matching. We report the matching precision (M.Precision), the matching recall (M.Recall), the median positional errors (MPE, no scale), and the median rotation errors (MRE, $^\circ$)}
  \begin{tabular}{lcccc}
  \toprule
  Matcher & M.Precision & M.Recall  & MPE & MRE\\
  \midrule
  NN+ratio & 52.5 & 28.6 & 0.493 & 0.044 \\
  NN+cross & 36.0 & 38.0 & 0.416 & 0.038 \\
  NN+distance & 36.7 & 40.7 & 0.382 & 0.036 \\
  \midrule
  NN+LGCNet-2D3D & 55.0 & 41.5 & 0.374 & 0.036 \\
  NN+OANet-2D3D & 55.8 & 42.3  & 0.364 & 0.036 \\
  \midrule
  GAM    & \textbf{57.0} & \textbf{50.8} & \textbf{0.335} & \textbf{0.030} \\
  \bottomrule
  \end{tabular}
  \label{tab:matcher}
\end{table}

Results are listed in Table \ref{tab:matcher}. Compared with other handcrafted methods, although NN+ratio has high matching precision, its pose error is the largest in all handcrafted methods. This is because a large number of correct matches are filtered after ratio test. Some qualitative 2D-3D matches are illustrated in Fig.\ref{fig:2D3Dmatching}. Due to the existence of repetitive patterns, NN+ratio shows few matches compared to other methods. Compared with NN+LGCNet-2D3D, NN+OANet-2D3D is better under various metrics. The improvement is mainly brought by the usage of the local context of 2D-3D correspondences. GAM has a significantly higher recall than both handcrafted and learned baselines, which can attribute to $k$NN ratio matching that can recall the real matches that are not in the nearest neighbors. Meanwhile, although GAM takes more ambiguous matches into consideration, it also maintains a higher precision than hand-crafted and learned methods. This is because BMNet equipped with the Hungarian pooling layer has the ability to find out true matches from original many-to-many ambiguous matches. Fig.\ref{fig:2D3Dmatching} also shows the effectiveness of the proposed matching method GAM.

\subsection{Visual Localization}
We continue to evaluate the effectiveness of GAM in visual localization and our proposed visual localization framework.

\textbf{Dataset.} We conduct experiments on Aachen Day-Night dataset\cite{sattler2012image},  which is a challenging large dataset introduced by \cite{sattler2018benchmarking}. 
% The images of Aachen Day-Night dataset consist of two parts, the reference images that are used to construct the sparse SfM model and the query images that are used for evaluation.  All the query images are collected using mobile phones and are annotated with the ground-truth 6DoF poses, which are very suitable for augmented reality scenes. The light changes during the day and night bring great challenges to visual localization.
The Aachen Day-Night dataset is composed of two distinct sets of images: reference images utilized for constructing the sparse SfM model and query images employed for evaluation. All query images are captured using mobile phones and are annotated with ground-truth 6DoF poses, rendering them highly suitable for augmented reality scenes. The alterations in lighting conditions between day and night present significant challenges for visual localization.
  
\textbf{Metric.} We report the pose recall at different accuracy levels of positions and orientations. We follow the benchmark \cite{sattler2018benchmarking} and use three accuracy levels ($0.25m,2^{\circ}$), ($0.5m,5^{\circ}$) and ($5.0m,10^{\circ}$).

\textbf{Feature Matching.} We first evaluate the effectiveness of GAM in visual localization. In this experiment, we intentionally just use the first scene provided by scene retrieval (SR$^1$) to perform feature matching. Our focus is on the performance of various matching methods in visual localization. We compare GAM with traditional hand-crafted methods, including NN+ratio, NN+distance, and NN+cross. We also compare our method with learned methods NN+LGCNet-2D3D and NN+OANet-2D3D as in the 2D-3D matching experiment. At the image retrieval stage, we use NetVLAD\cite{arandjelovic2016netvlad} to retrieve the top-50 images for all matching methods.

\begin{table}[htb!]
  \setlength{\tabcolsep}{1mm}
  \caption{
  % Visual Localization on the Aachen Day-Night datasets. We report the pose recall [$\%$] at different accuracy levels of positions ($m$) and orientations ($deg$).  
  Visual Localization on the Aachen Day-Night datasets. We report the pose recall [$\%$] at three accuracy levels: ($0.25m,2^{\circ}$), ($0.5m,5^{\circ}$) and ($5.0m,10^{\circ}$).
  \textbf{Bold} numbers denote the best result.}
  \centering %
  \begin{tabular}{llcc}
  \toprule
  \multicolumn{2}{c}{Method} & \multicolumn{2}{c}{Aachen Day-Night} \\
  \cmidrule(lr){1-2} 
  \cmidrule(lr){3-4} 
  Coarse & Fine & Day & Night \\
  \midrule
  \multirow{6}{*}{SR$^1$} & NN+ratio          & 79.7 / 88.8 / 93.7 & 54.1 / 70.4 / 81.6 \\ 
                          & NN+cross          & 83.7 / 90.9 / 94.1 & 68.4 / 78.6 / 85.7 \\
                          & NN+distance       & 82.0 / 90.2 / 94.1 & 66.3 / 78.6 / 84.7 \\
                      
                        & NN+LGCNet-2D3D    & 84.3 / 90.7 / \textbf{94.3} & 69.4 / 78.6 / \textbf{88.8} \\
                        & NN+OANet-2D3D     & 82.6 / 90.0 / 94.1 & 67.3 / 77.6 / 85.7 \\
                        & GAM               & \textbf{84.7 / 91.3 / 94.3} & \textbf{73.5 / 83.7 / 88.8} \\
  \midrule
  \midrule
  IR        & \multirow{3}{*}{GAM}          & 85.7 / 93.3 / 96.8 & 73.5 / 86.7 / 93.9  \\
  CC        &                               & 86.9 / 93.9 / 97.8 & 76.5 / 88.8 / 95.9  \\
  SR        &                               & \textbf{88.0 / 94.8 / 98.5} & \textbf{78.6 / 91.8 / 99.0}   \\
  \bottomrule
  \end{tabular}
  \label{tab:exp_loc}
\end{table}

\begin{figure}[htb!]
  \centering
  \includegraphics[width=1.0\linewidth]{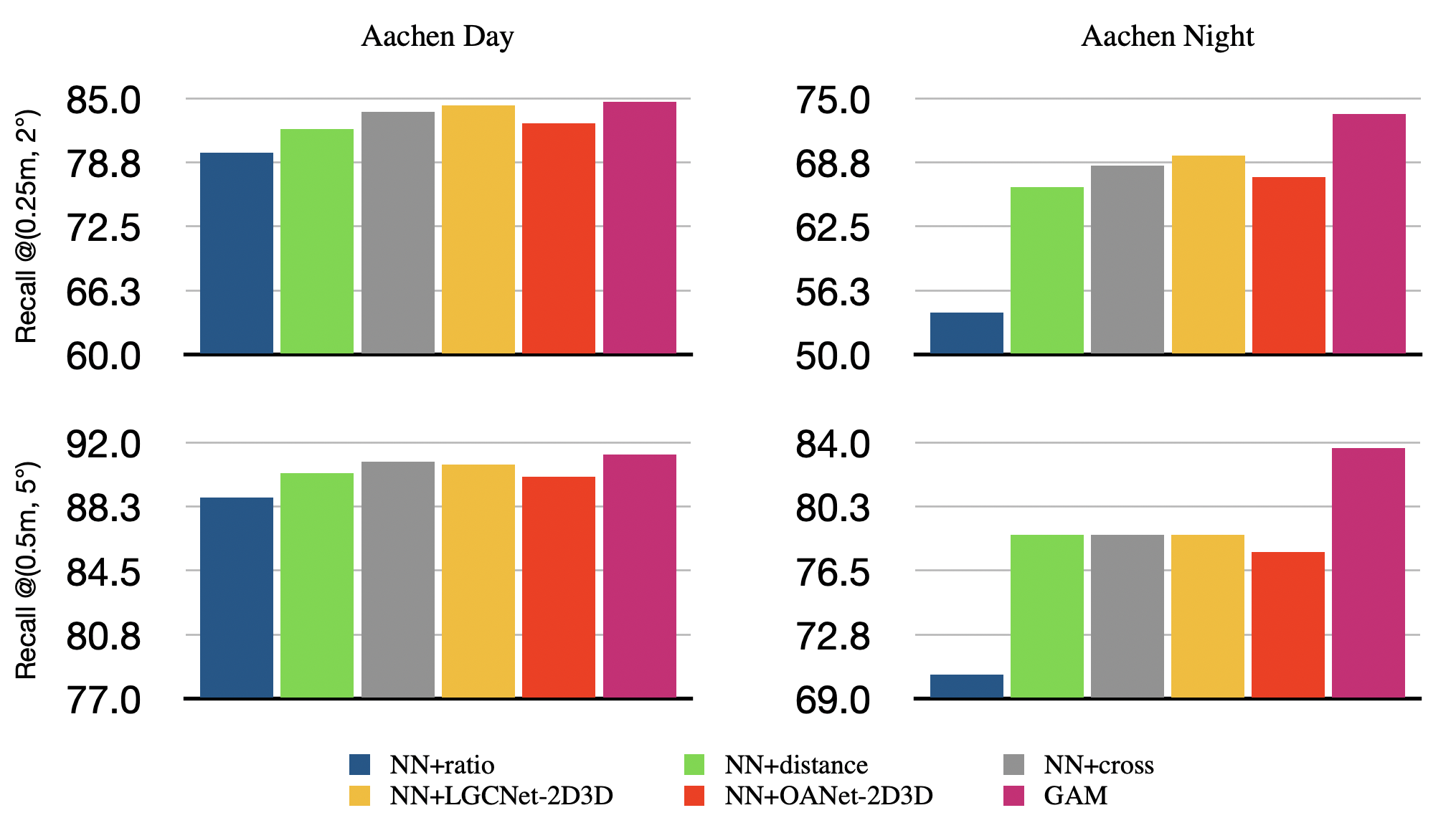}
  \caption{Pose recall under two high precision thresholds on Aachen Day-Night dataset. GAM consistently outperforms all baseline methods and shows a larger improvement on the night queries than the day. }
  \label{fig:vlmatch}
\end{figure}

\begin{table*}[!h]
  \setlength{\tabcolsep}{3mm}
  \caption{Visual localization on the Cambridge Landmarks dataset. We report the median translation ($cm$) and rotation ($\circ$) errors on five scenes and compute the mean errors. We mark best results \textbf{bold}.}
  \centering %
  \begin{tabular}{llcccccc}
  \toprule
 \multicolumn{2}{c}{Method}       & Great Court          & Kings College       & Old Hospital & Shop Facade  & St M. Church &  Avg\\
  \midrule
  \multirow{6}{*}{E2E} & PoseNet\cite{kendall2015posenet}      & 700cm, 3.7$^{\circ}$ & 99cm, 1.1$^{\circ}$ & 217cm, 2.9$^{\circ}$ & 105cm, 4.0$^{\circ}$ & 149cm, 3.4$^{\circ}$ & 254cm, 3.02$^{\circ}$  \\
  & DFNet \cite{chen2022dfnet} & -  & 43cm, 0.87$^{\circ}$ & 46cm, 0.87$^{\circ}$ & 16cm, 0.59$^{\circ}$ & 50cm, 1.49$^{\circ}$ & -  \\
  & NG-DASC \cite{brachmann2019neural}   & 34.8cm, 0.18$^{\circ}$ & 12.2cm,0.23$^{\circ}$ & 21.2cm, 0.45$^{\circ}$ & 5.4cm, 0.29$^{\circ}$ & 9.9cm, 0.31$^{\circ}$ & 16.7cm, 0.29$^{\circ}$ \\
  & DSAC$^\ast$ \cite{brachmann2021visual}       & 34.0cm, 0.2$^{\circ}$ & 18.0cm, 0.3$^{\circ}$ & 21.0cm, 0.4$^{\circ}$ & 5.0cm, 0.3$^{\circ}$ & 15.0cm, 0.6$^{\circ}$ & 18.6cm, 0.36$^{\circ}$ \\
  & PixLoc \cite{sarlin2021back}    & 30.0cm, 0.14$^{\circ}$ & 14.0cm,0.24$^{\circ}$ & 16.0cm, 0.32$^{\circ}$ & 5.0cm, 0.23$^{\circ}$ & 10.0cm, 0.34$^{\circ}$ & 15.0cm, 0.25$^{\circ}$\\
   & SC-wLS  \cite{wu2022sc}    & 29.0cm, 0.2$^{\circ}$ & 8.0cm, 0.20$^{\circ}$ & 11.0cm, 0.40$^{\circ}$ & 4.0cm, 0.30$^{\circ}$ & 9.0cm, 0.30$^{\circ}$ & 12.2cm, 0.28$^{\circ}$\\
  \midrule
  FM(2D2D) & HLoc+SG \cite{sarlin2020superglue}    & \textbf{10.1cm, 0.07}$^{\circ}$ & 6.9cm,0.11$^{\circ}$ & 12.5cm, 0.24$^{\circ}$ & 2.9cm, 0.14$^{\circ}$ & 3.8cm, 0.12$^{\circ}$ &  7.2cm, 0.14$^{\circ}$ \\
  \midrule
  
  \multirow{2}{*}{FM(2D3D)} & AS \cite{sattler2012improving}        & 24.0cm, 0.13$^{\circ}$ & 13.0cm,0.22$^{\circ}$ & 20.0cm, 0.36$^{\circ}$ & 4.0cm, 0.21$^{\circ}$ & 8.0cm, 0.25$^{\circ}$ & 13.8cm, 0.23$^{\circ}$ \\
  & Ours      & 10.6cm, 0.08$^{\bm{\circ}}$ & \textbf{5.4cm, 0.10}$^{\bm{\circ}}$ & \textbf{11.0cm, 0.22}$^{\bm{\circ}}$ & \textbf{2.5cm, 0.13}$^{\bm{\circ}}$ & \textbf{3.5cm, 0.11}$^{\bm{\circ}}$ & \textbf{6.6cm, 0.13}$^{\bm{\circ}}$ \\ 
  \bottomrule
  \end{tabular}
  \label{tab:cmp_cambridge}
\end{table*}

Results are presented in Table \ref{tab:exp_loc}. GAM consistently outperforms all hand-crafted and learned baselines on both day and night queries, except at threshold ($5.0m,10^{\circ}$), under which GAM shows the same result with NN+LGCNet-2D3D. This confirms that GAM can effectively improve the accuracy of feature-based localization. It is worth noting that the improvement brought by GAM on the night queries is greater than that on the day queries, which is shown in Fig. \ref{fig:vlmatch}. This is because the discriminative power of appearance is reduced under the condition of severe light changes. In this case, the 2D-3D correspondences established only by finding the nearest neighbor in the descriptor space are not enough to provide accurate camera poses. GAM can effectively alleviate this problem, which shows the advantage of GAM in complex conditions. It should be pointed out that OANet-2D3D performs slightly worse than LGCNet-2D3D. We argue that this might be due to 3D points obtained by scene retrieval in large-scale scenes are not friendly to capturing local geometric context.

\textbf{Scene Retrieval.} We proceed to the evaluation of the proposed visual localization pipeline based on GAM. Our focus is on the performance of various coarse localization methods. At the image retrieval stage, we use NetVLAD\cite{arandjelovic2016netvlad} to retrieve the top-50 images. We compare the proposed scene retrieval (SR) with co-visible cluster\cite{sarlin2018leveraging} (CC). At the same time, we directly perform feature matching between the query features and 3D points provided by all retrieved images.  We mark this plain method as IR.

Results are shown in Table \ref{tab:exp_loc}. SR consistently outperforms both IR and CC. This comparison verifies the effectiveness of scene retrieval. The first reason for the improvement brought by scene retrieval is that it provides a cleaner 3D points set that is friendly to capture geometric context for GAM. The second is that scene retrieval is able to recall 3D points that would otherwise be lost due to image retrieval.

\begin{table*}[htb!]
  \setlength{\tabcolsep}{5mm}
  \caption{
  % Visual Localization on the Aachen Day-Night and RobotCar Seasons datasets. We report the pose recall [$\%$] at different accuracy levels of positions ($m$) and orientations ($deg$).  
  Visual Localization on the Aachen Day-Night and RobotCar Seasons datasets. We report the pose recall at three accuracy levels for all queries: ($0.25m,2^{\circ}$), ($0.5m,5^{\circ}$) and ($5.0m,10^{\circ}$).
  \textbf{Bold} numbers denote the best result.  A dash (-) indicates that the result was not reported by the corresponding methods.}
  \centering %
  \begin{tabular}{llcccc}
  \toprule
  \multicolumn{2}{c}{Method} & \multicolumn{2}{c}{Aachen Day-Night} & \multicolumn{2}{c}{RobotCar Seasons} \\
  \cmidrule(lr){1-2} 
  \cmidrule(lr){3-4} 
  \cmidrule(lr){5-6}
  Category & Name & Day & Night & Day & Night \\
  % &&($0.25m,2^{\circ}$)/($0.5m,5^{\circ}$)/($5.0m,10^{\circ}$)&($0.25m,2^{\circ}$)/($0.5m,5^{\circ}$)/($5.0m,10^{\circ}$) &
  % ($0.25m,2^{\circ}$)/($0.5m,5^{\circ}$)/($5.0m,10^{\circ}$) &
  % ($0.25m,2^{\circ}$)/($0.5m,5^{\circ}$)/($5.0m,10^{\circ}$) &
  % ($0.25m,2^{\circ}$)/($0.5m,5^{\circ}$)/($5.0m,10^{\circ}$) \\
  \midrule
  \multirow{2}{*}{E2E} &  ESAC \cite{brachmann2019expert} & 42.6 / 59.6 / 75.5 & 6.1 / 10.2 / 18.4 & - & - \\
  & PixLoc \cite{sarlin2021back} & 84.6 / 92.4 / 98.2 & 69.4 / 87.8 / 95.9 & 56.8 / 81.4 / \textbf{98.6}  & 8.8 / 25.6 / 58.2 \\
  \midrule
  FM(2D2D) & HLoc+SG \cite{sarlin2020superglue}  & \textbf{89.6 / 95.4 / 98.8} & \textbf{86.7 / 93.9 / 100.0} & 56.9 / 81.7 / 98.1 & \textbf{33.3 / 65.9 / 88.8}\\
  \midrule
  \multirow{4}{*}{FM(2D3D)} 
  & AS \cite{sattler2012improving} & 85.3 / 92.2 / 97.9 & 39.8 / 49.0 / 64.3 & 43.6 / 76.0 / 94.0  & 1.8 / 7.4 / 14.2 \\
  & CSL \cite{svarm2016city} & 52.3 / 80.0 / 94.3 & 29.6 / 40.8 / 56.1 & 45.3 / 73.5 / 90.1  & 0.6 / 2.6 / 7.2 \\
  & HLoc \cite{sarlin2019coarse} & 80.5 / 87.4 / 94.2 & 68.4 / 77.6 / 88.8 & 53.1 / 79.1 / 95.5  & 7.2 / 17.4 / 34.4 \\
  & Ours  & 88.0 / 94.8 / 98.5 & 78.6 / 91.8 / 99.0 & \textbf{57.8} / 81.6 / 97.4 & 12.6 / 35.3 / 69.4 \\
  \bottomrule
  \end{tabular}
  \label{tab:cmp_benchmark}
\end{table*}

\subsection{Comparison with State-of-the-art Methods}
We now compare our proposed pipeline with state-of-the-art methods on both the Cambridge Landmarks dataset and a large-scale long-term localization benchmark. 

\subsubsection{Cambridge Landmarks}
% Cambridge Landmarks dataset\cite{kendall2015posenet} is commonly used for end-to-end learning methods and contains six medium-scale outdoor scenes. Each scene includes the training images and the test images which are collected on different paths and in different conditions. All images have the ground-truth camera poses obtained by SfM.
The Cambridge Landmarks dataset \cite{kendall2015posenet} is frequently employed for end-to-end learning methods and comprises six medium-scale outdoor scenes. Each scene features training and test images collected on different paths and under varying conditions. All images are equipped with ground-truth camera poses obtained via SfM.
The Street scene is excluded from evaluation due to the errors in the provided trajectory. The remaining five scenes are re-triangulated with SuperPoint by HLoc toolbox\cite{sarlin2020hloc}.

We compare our proposed pipeline to end-to-end methods (E2E), including PoseNet \cite{kendall2015posenet}, DFNet \cite{chen2022dfnet}, NG-DSAC \cite{brachmann2019neural}, DSAC$^\ast$ \cite{brachmann2021visual}, PixLoc \cite{sarlin2021back} and SC-wLS \cite{wu2022sc}.We also compare our method with state-of-the-art feature-based localization methods (FM). Active Search (AS) \cite{sattler2012improving} directly perform feature matching between 2D and 3D (2D3D). HLoc+SuperGlue (HLoc+SG) \cite{sarlin2019coarse, sarlin2020superglue}, a state-of-the-art 2D-2D localization method, establishes 2D-3D correspondences by performing 2D-2D feature matching many times (2D2D). We use NetVLAD to retrieve top-10 reference images for HLoc+SG and ours.

Results are presented in Table \ref{tab:cmp_cambridge}. 
% The NG-DSAC results are obtained by using the model released by the author. The results of PixLoc and AS come from \cite{sarlin2021back}. 
The result of HLoc+SG  is obtained from our implementation based on the released model by the author.  We found that our re-implemented result is better than the author's. 
% As can be seen from Table \ref{tab:cmp_cambridge}, the methods that regress scene coordinates, such as NG-DSAC, DSAC$^\ast$, and SC-wLS, perform better than the methods that directly regress camera pose, such as PoseNet and DFNet. Our method is more accurate than the methods of regressing scene coordinates and PixLoc. These results indicate that for the complicated task of localization, using the learning in the specific modules may be more effective than learning the whole process.
Table \ref{tab:cmp_cambridge} demonstrates that the methods, such as NG-DSAC, DSAC$^\ast$, and SC-wLS, that regress scene coordinates exhibit better performance than the methods that directly regress camera pose, such as PoseNet and DFNet. Additionally, our method proves to be more accurate than the methods of regressing scene coordinates and PixLoc. These findings suggest that for the challenging task of localization, utilizing learning in specific modules may be more effective than learning the entire process.
Not only that, but our method also consistently outperforms AS by a large margin in all scenes. Compared with the state-of-the-art 2D-2D localization method, our method outperforms HLoc+SG in all scenes except Great Court.  Note that although HLoc+SG has higher accuracy than our method on Great Court, its speed is about $6$ times slower than our method, which can be seen in Fig.\ref{fig:cmp_time}.

\subsubsection{Large-scale Localization}
% We compare our method with state-of-the-art methods on the Aachen Day-Night dataset and the RobotCar Seasons dataset. The RobotCar dataset \cite{maddern20171} consists of several video sequences collected in different seasons, including 26,121 reference images and 11,934 query images. The reference images from one season are used to construct the SfM model. Compared to Aachen Day-Night, the query images are even more challenging because both seasonal changes and illumination changes are included.
Our method is compared with state-of-the-art methods on two datasets: the Aachen Day-Night dataset and the RobotCar Seasons dataset. The RobotCar dataset \cite{maddern20171} comprises various video sequences captured in different seasons, including 26,121 reference images and 11,934 query images. The reference images from a single season are employed to construct the SfM model. Unlike Aachen Day-Night, the query images in this dataset are even more challenging due to the inclusion of both seasonal and illumination changes.

We compare our method with Active Search (AS) \cite{sattler2012improving}, City Scale Localization (CSL) \cite{svarm2016city}, HLoc \cite{sarlin2019coarse} and  HLoc+SuperGlue (HLoc+SG) \cite{sarlin2019coarse, sarlin2020superglue}. These methods are based on feature matching (FM). In addition, we compare against end-to-end learning-based methods (E2E), ESAC \cite{brachmann2019expert} and PixLoc\cite{sarlin2021back}.
We use the results retrieved by NetVLAD, which is the same as HLoc \cite{sarlin2019coarse}. The evaluation metric follows the benchmark\cite{sattler2018benchmarking}. 

Results are shown in Table \ref{tab:cmp_benchmark}. Our method consistently outperforms ESAC and outperforms the recently proposed PixLoc except on the day queries of RobotCar under the threshold ($5m,10^{\circ}$). On two datasets, our method also consistently outperforms localization methods based on 2D-3D matching including AS, CSL, and HLoc.  It is worth mentioning that every 3D point used in HLoc is represented by all 2D descriptors in the track when performing 2D-3D feature matching, while our method only uses the mean of these descriptors, which greatly reduces the requirement of storage and timing. This confirms that our method is effective. Compared with the state-of-the-art 2D-2D localization method, although HLoc+SG establishes 2D-3D correspondences by performing 2D-2D feature matching many times, our method performs comparably to HLoc+SG on the Aachen Day-Night dataset and the day queries of RobotCar Seasons dataset. Note that our method is slightly better than HLoc+SG on the day queries of RobotCar Seasons dataset under threshold ($0.25m, 2^{\circ}$).

\subsection{Ablation Study}
We will make ablation studies about our proposed method in this section. We present experiments on the 2D-3D matching dataset. Our focus is on the behaviors of GAM with different configurations. 

\begin{figure}[htb!]
  \centering
  \includegraphics[width=1\linewidth]{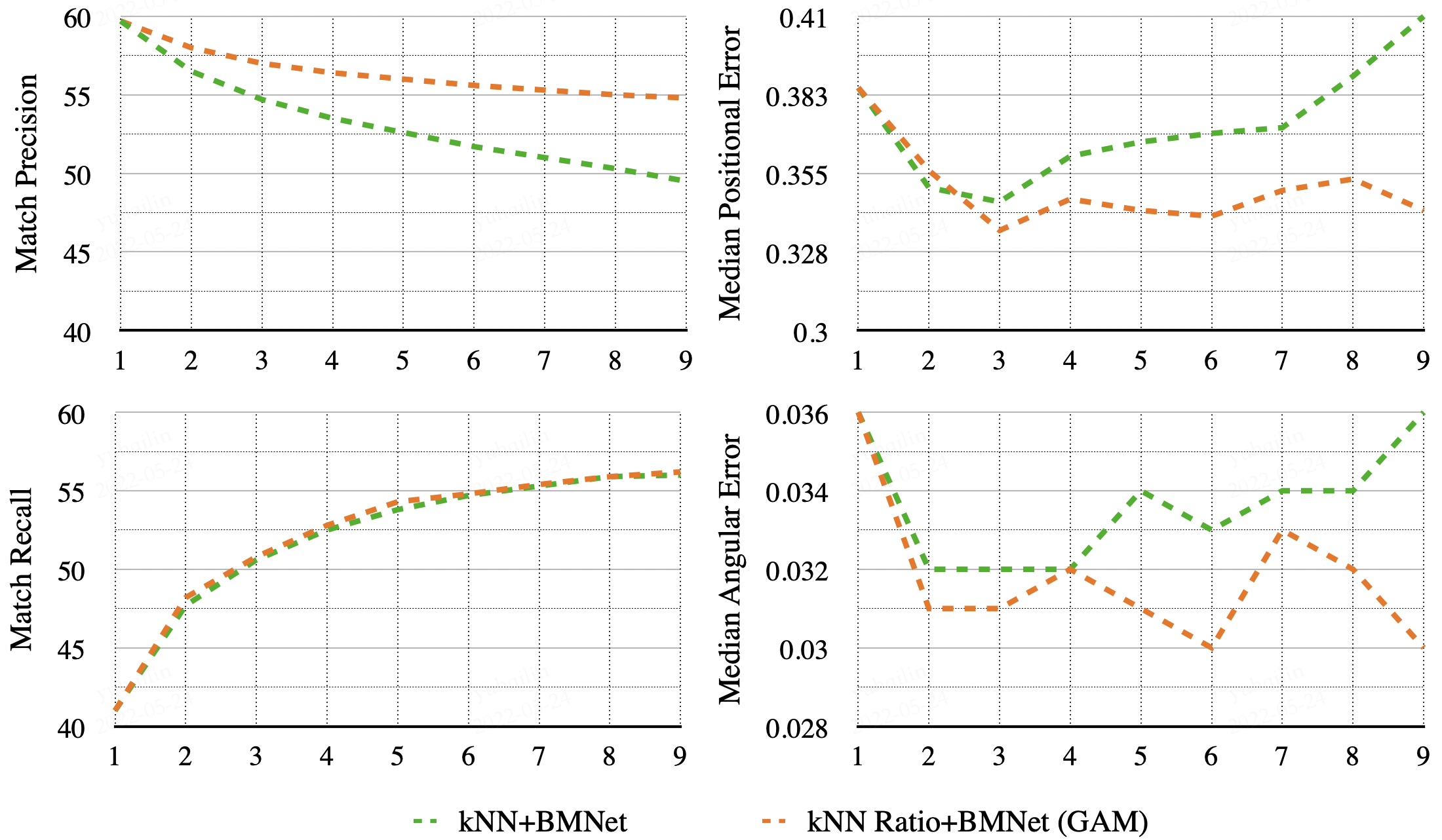}
  \caption{Match precision (top left), match recall (bottom left), median positional error (top right), and median angular error (bottom right) on the test dataset. Horizontal coordinate represents hyper-parameter $k$.} 
  \label{fig:gamk}
\end{figure}

\textbf{k-Nearest Neighbors.} We evaluate the impact of hyper-parameter $k$ in GAM. The ratio threshold is fixed as 0.7, aiming to filter obviously false matches. We have two significant observations from Fig.\ref{fig:gamk}. First, with the increase of $k$, the recall increases, and the precision slightly decreases. This indicates that GAM can effectively strengthen the recall of 2D-3D matches while maintaining high precision. Second, with the increase of $k$, both MPE and MAE are dropped first. The median error reaches the minimum at $k=3$, which is improved by about 13\% compared to $k=1$.  When $k$ increases, both MPE and MAE fluctuate, but the fluctuation range is small and will not perform worse than when $k=1$. This verifies the effectiveness of leveraging $k$ nearest neighbors. 

\textbf{kNN vs. kNN Ratio.}
To demonstrate the effectiveness of $k$NN ratio,  we compare $k$NN ratio+BMNet (GAM) with $k$NN+BMNet under various $k$. The only difference between both methods is the candidate matches fed into BMNet.  The results are presented in Fig.\ref{fig:gamk}.  First,  the match recall of both $k$NN ratio+BMNet and $k$NN+BMNet are almost identical, but $k$NN+BMNet has lower match precision than $k$NN ratio+BMNet. Both the minimum MAP and MPE of $k$NN+BMNet are larger than that of $k$NN ratio+BMNet.  Second, with the increase of $k$, both MPE and MAE of $k$NN+BMNet are dropped first and then increased,  which is different from $k$NN ratio+BMNet which shows fluctuation after reaching the minimum.  The MPE of $k$NN+BMNet at $k=8$ even exceeds that at $k=1$. This indicates that $k$NN ratio+BMNet is less sensitive to $k$ value than $k$NN+BMNet and shows more robustness.

\textbf{Hungarian Pooling.}
We train a model without Hungarian pooling (HP) denoted as PlainNet. We compare PlainNet and PlainNet+HP with BMNet. Hungarian pooling layer is parameter-free,  so PlainNet has the same number of parameters as BMNet.  The initial candidate 2D-3D matches are generated by $k$NN ratio matcher for both models.  As shown in Table \ref{tab:cmp_component},  although PlainNet has a higher matching recall than BMNet,  it shows lower matching precision and poses accuracy. This indicates that Hungarian pooling is able to filter out geometrically ambiguous matches, which have a negative effect on pose estimation. BMNet also outperforms PlainNet+HP.  The only difference between the two methods is that BMNet is trained end-to-end with Hungarian pooling.  This comparison verifies the effectiveness of end-to-end training with Hungarian pooling layer.

\begin{table}[htb!]
  \setlength{\tabcolsep}{2mm}
  \centering %
  \caption{Ablation Study. We report the matching precision (M.Precision), the matching recall (M.Recall), the median positional errors (MPE, no scale), and the median rotation errors (MRE, $^\circ$). The candidate matches are obtained by kNN ratio}
  \begin{tabular}{lcccc}
  \toprule
  Matcher & M.Precision & M.Recall  & MPE & MRE \\
  \midrule
  PlainNet            & 16.7 & 62.8 & 0.471 & 0.044 \\
  PlainNet+HP         & 53.6 & 49.2 & 0.372 & 0.036 \\
  \midrule
  BMNet (w/o mining)  & 55.2 & 50.1 & 0.342 & 0.032 \\
  \midrule
  BMNet               & 57.0 & 50.8 & 0.335 & 0.030 \\
  \bottomrule
  \end{tabular}
  \label{tab:cmp_component}
\end{table}

\textbf{Negative Sample Mining.}
We train a model without negative sample mining (mining) denoted as BMNet (w/o mining). When training BMNet (w/o mining), the negative samples are generated by the random selection that is adopted by \cite{yu2020learning}. The other configurations are kept the same as BMNet. We compare BMNet (w/o mining)  with BMNet.  The initial candidate 2D-3D matches are obtained by  $k$NN ratio matcher. As shown in Table \ref{tab:cmp_component}, BMNet outperforms BMNet (w/o mining) under all metrics.   This comparison verifies the effectiveness of negative sample mining at the training stage.

\subsection{Timing} 
\begin{figure}[htb!]
  \centering
  \includegraphics[width=1.0\linewidth]{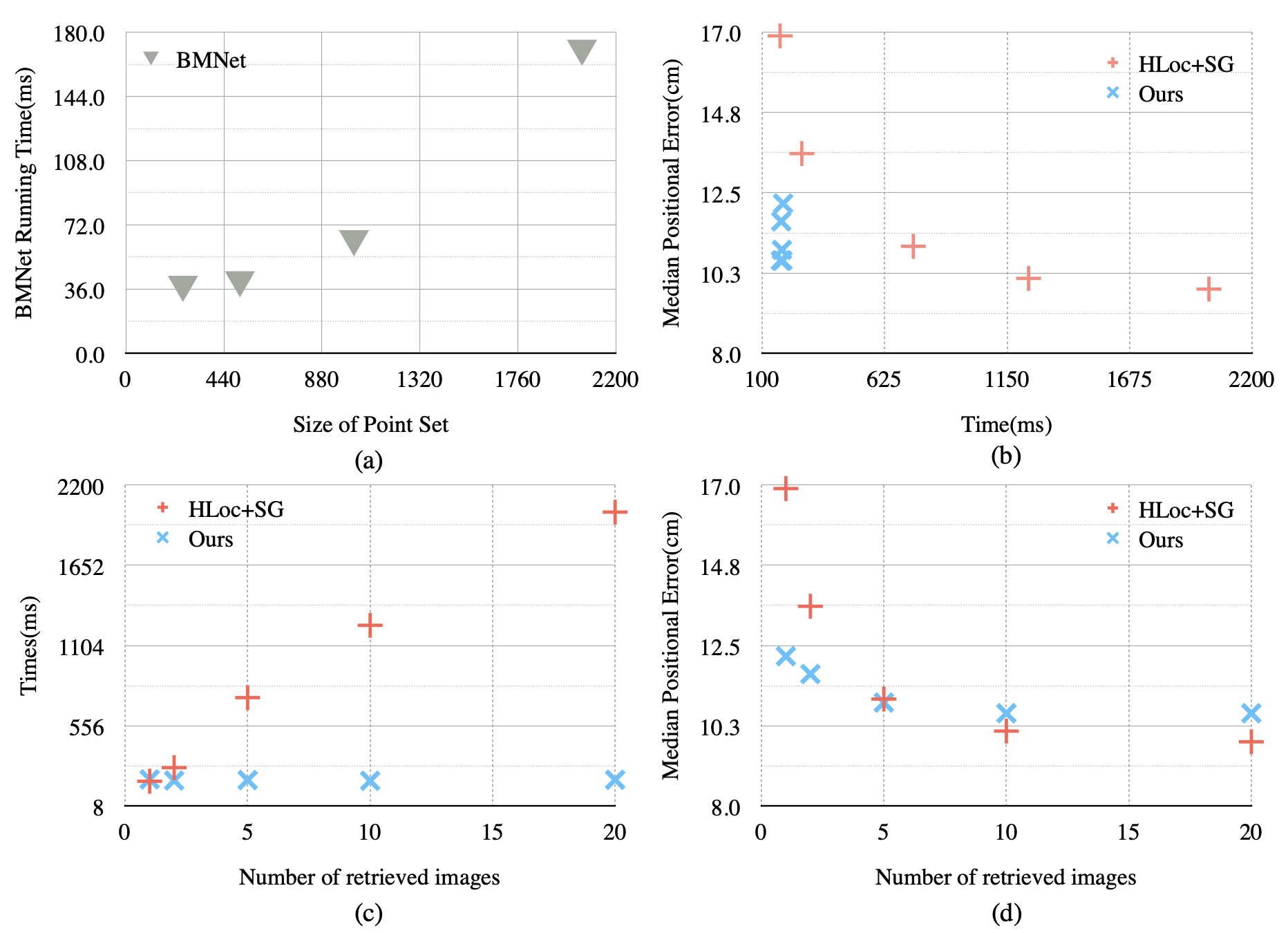}
  \caption{(a) Running time of BMNet with various sizes of the point set.  (b) The relationship between median positional error and elapsed time with various numbers of retrieved images. (c) The relationship between elapsed time and the number of retrieved images. (d) The relationship between median positional error and the number of retrieved images.}
  \label{fig:cmp_time}
\end{figure}

% We measure the running time of the main components of the proposed method on the machine with an Intel Core i7-8700 CPU and a GeForce GTX 1080 GPU.
The running time of the primary components of the proposed method is measured on a machine equipped with an Intel Core i7-8700 CPU and a GeForce GTX 1080 GPU.
We resize the larger dimension of the query images to 1024 for SuperPoint and NetVLAD. We calculate the average running time on the Great Court scene of Cambridge Landmarks dataset \cite{kendall2015posenet}. The scene includes 153,2 reference images and 760 query images. The run-time of main components are presented in Table \ref{tab:time_component}.  The average number of processed scenes on this dataset is $1.06$, so the average time of the whole pipeline is $309.59$ms. 
% The details about the running time of BMNet used in GAM can be seen in Fig.\ref{fig:cmp_time}(a). The sizes of the 2D point set and 3D point set are equal and the size of the edge set is twice as much as the size of the 2D point set. BMNet takes 39.39ms and 62.4ms when the point set size is 512 and 1024, respectively.
The running time details of BMNet utilized in GAM are illustrated in Fig.\ref{fig:cmp_time}(a). The sizes of the 2D and 3D point sets are equivalent, while the edge set size is double that of the 2D point set. For point set sizes of 512 and 1024, BMNet requires 39.39ms and 62.4ms, respectively.

\begin{table}[htb!]
  \setlength{\tabcolsep}{3mm}
  \centering %
  \caption{Running Time. We report the mean run-time (ms) of different components of the proposed localization method. SP and NV represent SuperPoint and NetVLAD. SR represents scene retrieval. PE represents prior-guided pose estimation.}
  \begin{tabular}{lccccc}
  \toprule
  Component & SP & NV & SR & GAM & PE\\
  \midrule
  Times(ms)   & 35.33 & 80.62 & 14.62 & 117.43 & 50.63 \\
  \bottomrule
\end{tabular}
\label{tab:time_component}
\end{table}

\textbf{The Number of Retrieved Images.} We conduct more studies about the relationship between the median positional error and running time with the number of retrieved images. We present the results of our method and HLoc+SG \cite{sarlin2020superglue} in Fig.\ref{fig:cmp_time}(b)(c)(d). Note the time of extracting SuperPoint and NetVLAD is excluded here. The elapsed time of our method is insensitive to the number of retrieved images, while the elapsed time of HLoc+SG is linearly related to the number of retrieved images. In addition, HLoc+SG is more dependent on the number of retrieved images than our method in terms of pose accuracy. Although HLoc+SG is slightly accurate than ours as the number of retrieved images increases to 10, it takes more than six times as long as ours.

\subsection{Application for Augmented Reality}
Global localization with 6DoF pose estimation is crucial for augmented reality applications in a large-scale scene. In this section, we use the proposed visual localization method to make an AR application.
 
% We capture a video in a scene covering about 20,000 square meters. 
We capture a video in an office covering about 1,000 square meters. We extract frames from this video as reference images that are used to reconstruct a sparse SfM model with SuperPoint. Then we choose some locations to place some virtual 3D arrows. The coordinates of these virtual objects are aligned to the SfM model.
 
\begin{figure}[htb!]
  \centering
  \includegraphics[width=1.0\linewidth]{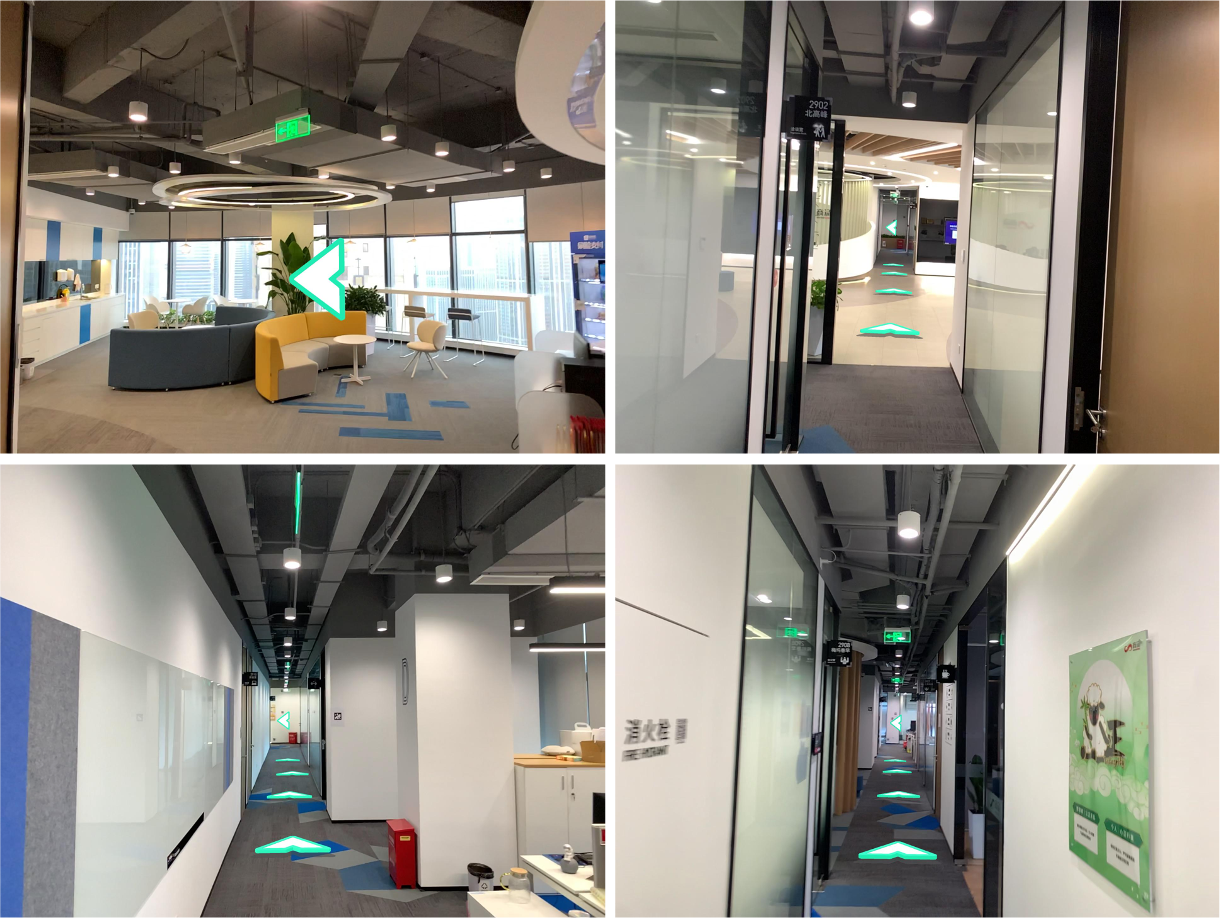}
  \caption{Four selected augmented frames. The arrows on the ground and in the air are pre-aligned virtual objects.}
  \label{fig:AR}
\end{figure}

\begin{figure}[htb!]
  \centering
  \includegraphics[width=1.0\linewidth]{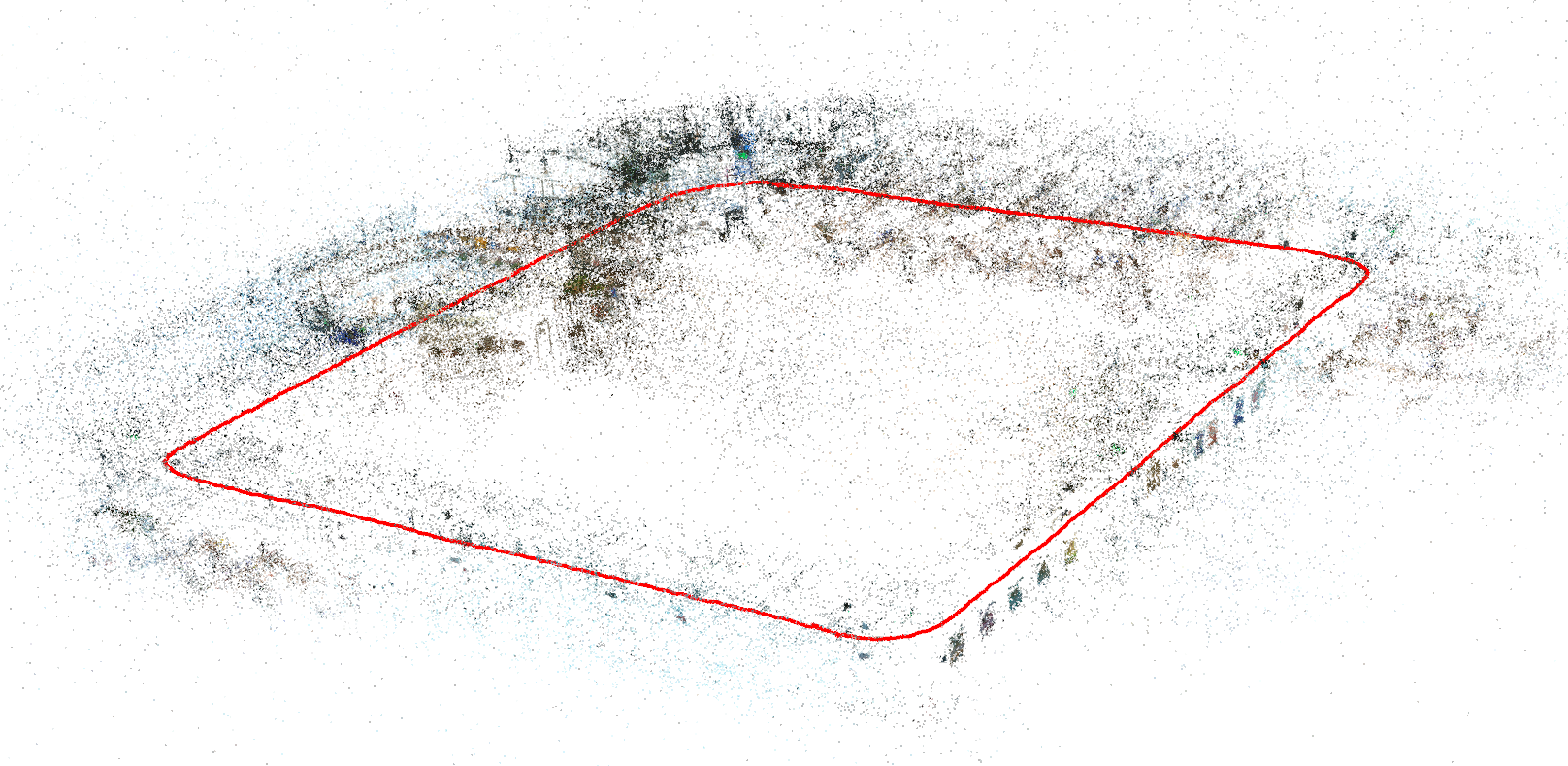}
  \caption{The recovered camera trajectory superimposed in the offline 3D map recovered by SfM. All poses are faithfully recovered.}
  \label{fig:Loc_traj}
\end{figure}

We capture a query video and extract frames from the query video at 30 fps. To demonstrate the accuracy and robustness of our localization result, we use the proposed localization method to recover the camera poses for all frames. Then we render the pre-aligned virtual objects on these frames according to the localization results. Some AR frames are shown in Fig.~\ref{fig:AR}.  The recovered camera trajectory is shown in Fig.~\ref{fig:Loc_traj}. 
% It can be seen that the recovered trajectory by our method is already quite smooth even though the pose of each frame is estimated independently. All the poses of all frames are faithfully recovered, which demonstrates the effectiveness of the proposed visual localization method. Please refer to the supplementary video for watching the whole result.
It is evident that the recovered trajectory by our method is already quite smooth despite the fact that the pose of each frame is estimated independently. All the frame poses are accurately recovered, which shows the efficacy of the proposed visual localization method. Please refer to the supplementary video for viewing the entire result.

% Actually, for AR applications, we do not need to estimate the pose of each frame independently by global relocalization, since SLAM  ~\cite{klein2007parallel, mur2015orb}  technique can be used to smoothly recover the poses of each online frame. Global relocalization can be used to align the 3D coordinate of SLAM into the world coordinate and correct the pose to eliminate tracking drift. 
In actuality, we do not need to estimate the pose of each frame independently through global relocalization for AR applications, as SLAM  ~\cite{klein2007parallel, mur2015orb}  technique can be employed to smoothly recover the poses of each online frame. Global visual relocalization can be utilized to align the 3D coordinate of SLAM with the world coordinate and adjust the pose to eliminate tracking drift.

 In this application, we use ARkit\footnote{https://developer.apple.com/augmented-reality/arkit} to recover the camera pose in real-time. We first use the proposed localization to align ARkit results to SfM coordinate. Then the localization and alignment are performed every 10 seconds. Please refer to the supplementary video for watching the AR effect. Since our main focus is on visual localization, we use this simple strategy to couple SLAM and global localization results. Some more effective couple methods can be referred to \cite{lynen2015get,zuo2019visual, yamaguchi2020global, BaoXQCZWZ22}.

\section{Conclusion}\label{conclusion}
% This paper proposes an effective 2D-3D matching method GAM for visual localization. GAM uses both the appearance information and geometric context to improve the matching performance, which greatly improves the recall of 2D-3D matching while maintaining high precision. GAM introduces a novel bipartite matching neural network BMNet to extract geometric features for a set of 2D-3D correspondences and can learn the global geometric consistency to predict the possibility of being a true match for each correspondence. GAM also integrates the Hungarian algorithm into BMNet as a special pooling layer to find maximum-weight matching in an end-to-end manner. This approach enables the localization to obtain more correct matches and hence improves the robustness and accuracy of localization. We further combine GAM with a novel scene retrieval strategy and propose a new hierarchical localization method. Extensive experiments show that the proposed method achieves state-of-the-art results on multiple datasets. 
In this work, we introduce GAM, a reliable and efficient 2D-3D matching technique for visual localization, which significantly boosts the recall of 2D-3D matches while preserving high accuracy by leveraging both appearance information and geometric context to enhance matching performance.  GAM also introduces a novel bipartite matching neural network BMNet to identify the maximum weight matching in an end-to-end fashion, which allows the localization method to acquire more real matches.  This further increases the robustness and accuracy of localization. Additionally, we combine GAM with a novel scene retrieval strategy and present a new hierarchical localization method. Extensive experiments demonstrate that the proposed method achieves superior results on multiple datasets.

\section*{Acknowledgments}
This work was partially supported by NSF of China (Nos.61932003 and 61822310).

% {\appendix[Proof of the Zonklar Equations]
% Use $\backslash${\tt{appendix}} if you have a single appendix:
% Do not use $\backslash${\tt{section}} anymore after $\backslash${\tt{appendix}}, only $\backslash${\tt{section*}}.
% If you have multiple appendixes use $\backslash${\tt{appendices}} then use $\backslash${\tt{section}} to start each appendix.
% You must declare a $\backslash${\tt{section}} before using any $\backslash${\tt{subsection}} or using $\backslash${\tt{label}} ($\backslash${\tt{appendices}} by itself
%  starts a section numbered zero.)}

 \bibliographystyle{IEEEtran}
% \bibliography{IEEEabrv,../bib/paper}
\bibliography{IEEEabrv, main}

\vfill

\end{document}